\documentclass[10pt,journal,compsoc]{IEEEtran}
\usepackage[utf8]{inputenc}
\usepackage[nocompress]{cite}
\usepackage[pdftex]{graphicx}
\usepackage{amsmath}
\usepackage{amssymb}
\usepackage{amsfonts}
\usepackage{dsfont}
\usepackage{stmaryrd}
\usepackage[caption=false,font=footnotesize,labelfont=sf,textfont=sf]{subfig}
\usepackage[export]{adjustbox}
\usepackage[usenames, dvipsnames]{color}
\usepackage{algorithmic}

\newcommand{\norm}[1]{\left\lVert#1\right\rVert}
\newcommand{\parent}[1]{\left(#1\right)}

\newcommand{\dotprod}[1]{\left<#1\right>}
\renewcommand{\brack}[1]{\left[#1\right]}
\renewcommand{\brace}[1]{\left\{#1\right\}}
\renewcommand{\det}[1]{\left|#1\right|}

\newcommand{\crossent}[2]{H\parent{#1,#2}}

\newcommand{\E}[1]{\mathbb{E}\brack{#1}}
\newcommand{\argmin}[1]{\underset{#1}{\text{argmin}} \,}
\newcommand{\argmax}[1]{\underset{#1}{\text{argmax}} \,}

\renewcommand{\P}[1]{\mathbb{P}\parent{#1}}

\renewcommand{\max}[1]{\underset{#1}{\text{max}} \,}

\newcommand{\N}[1]{\mathcal{N}\parent{#1}}
\newcommand{\W}[1]{\mathcal{W}_p\parent{#1}}
\newcommand{\R}{\mathbb{R}}
\newcommand{\M}{\mathcal{M}}
\newcommand{\G}{\mathcal{G}}

\newtheorem{lemma}{Lemma}

\newenvironment{proposition}[1][{\textit{Proposition}:}]{\begin{trivlist}
\item[\hskip \labelsep {\bfseries #1}]}{\end{trivlist}}

\newenvironment{remark}[1][Remark]{\begin{trivlist}
\item[\hskip \labelsep {\bfseries #1}]}{\end{trivlist}}

\newcommand\independent{\protect\mathpalette{\protect\independenT}{\perp}}
\def\independenT#1#2{\mathrel{\rlap{$#1#2$}\mkern2mu{#1#2}}}

\begin{document}
\title{Gaussian Graphical Model exploration and selection in high dimension low sample size setting
}

%


\author{Thomas~Lartigue, Simona~Bottani, St\'ephanie~Baron, Olivier~Colliot, Stanley~Durrleman, St\'ephanie~Allassonni\`ere for the Alzheimer’s Disease Neuroimaging Initiative
\IEEEcompsocitemizethanks{\IEEEcompsocthanksitem T. Lartigue is with the 
CMAP, CNRS, \'Ecole polytechnique and Aramis project-team, Inria. E-mail: thomas.lartigue@inria.fr
\IEEEcompsocthanksitem S. Bottani is with the Aramis project-team, Inria, Institut du Cerveau et de la Moelle \'epini\`ere, Sorbonne University, Inserm U1127, CNRS UMR 7225. E-mail: simona.bottani@icm-institute.org
\IEEEcompsocthanksitem S. Baron is with H\^{o}pital Europ\'een Georges-Pompidou AP-HP. E-mail: stephanie.baron@aphp.fr
\IEEEcompsocthanksitem O. Colliot and S. Durrleman are with the Aramis project-team, Inria, Institut du Cerveau et de la Moelle \'epini\`ere, Sorbonne University, Inserm U1127, CNRS UMR 7225.\protect\\ E-mail: olivier.colliot@upmc.fr \& stanley.durrleman@inria.fr 
\IEEEcompsocthanksitem S. Allassonni\`ere is with the Centre de Recherche des Cordeliers, Universit\'e de Paris, INSERM, Sorbonne Universit\'e. E-mail: stephanie.allassonniere@parisdescartes.fr
\IEEEcompsocthanksitem Data used in the preparation of this article were obtained from the Alzheimer's Disease Neuroimaging Initiative (ADNI) database (http://www.loni.ucla.edu/ADNI). As such, the investigators within the ADNI contributed to the design and implementation of ADNI and/or provided data but did not participate in analysis or writing of this report. A complete listing of ADNI investigators can be found at https://adni.loni.usc.edu/wp-content/uploads/how\textunderscore to\textunderscore apply/ADNI\textunderscore Acknowledgement\textunderscore List.pdf}
}


\IEEEtitleabstractindextext{%
\begin{abstract}
Gaussian Graphical Models (GGM) are often used to describe the conditional correlations between the components of a random vector. In this article, we compare two families of GGM inference methods: the nodewise approach of \cite{meinshausen2006high} and \cite{giraud2012graph} and the penalised likelihood maximisation of \cite{yuan2007model} and \cite{banerjee2008model}. We demonstrate on synthetic data that, when the sample size is small, the two methods produce graphs with either too few or too many edges when compared to the real one. As a result, we propose a composite procedure that explores a family of graphs with a nodewise numerical scheme and selects a candidate among them with an overall likelihood criterion. We demonstrate that, when the number of observations is small, this selection method yields graphs closer to the truth and corresponding to distributions with better KL divergence with regards to the real distribution than the other two. Finally, we show the interest of our algorithm on two concrete cases: first on brain imaging data, then on biological nephrology data. In both cases our results are more in line with current knowledge in each field. 
\end{abstract}

}

\maketitle

\IEEEdisplaynontitleabstractindextext

%
\IEEEpeerreviewmaketitle

\ifCLASSOPTIONcompsoc
\IEEEraisesectionheading{\section{Introduction}\label{sec:introduction}}
\else
\section{Introduction}
\label{sec:introduction}
\fi

%
%
%
%
\IEEEPARstart{D}{ependency} networks are a prominent tool for the representation and interpretation of many data types as, for example, gene co-expression \cite{giraud2012graph}, interactions between different regions of the cortex \cite{bullmore2009complex} or population dynamics. In those examples, the number of observations $n$ is often small when compared to the number of vertices $p$ in the network.\\
Conditional correlation networks are graphs where there exists an edge between two vertices if and only if the random variables on these nodes are correlated conditionally to all others. This structure can be more interesting than a regular correlation graph. Indeed, in real life, two phenomena, like the atrophy in two separate areas of the brain or two locations of bird migration, are very likely to be correlated. There almost always exists a "chain" of correlated events that "link", ever so slightly, any two occurrences. As a result, regular correlation networks tend to be fully connected and mostly uninformative. On the other hand, when intermediary variables explain the totality of the co-variations of two vertices, then these two are conditionally uncorrelated, removing their edge from the conditional correlation graph. The conditional correlation structure captures only the direct, explicit interactions between vertices. In our analyses, these interactions are the ones of most interest.\\
\\
A Gaussian Graphical Model (GGM) is a network whose values on the $p$ vertices follow a Centred Multivariate Normal distribution in $\R^p$: $X \sim \N{0_p, \Sigma}$. This assumption is almost systematic when studying conditional correlation networks for three main reasons. First, it ensures that each conditional correlation $corr(X_i, X_j | (X_k)_{k \neq i,j})$ is a constant and not a function of the $p-2$ dimensional variable $(X_k)_{k \neq i,j}$; a crucial property allowing us to talk about a single graph and not a function graph. Second, it equates the notions of independence and un-correlation, in particular: $corr(X_i, X_j | (X_k)_{k \neq i,j}) = 0 \iff  X_i \independent X_j| (X_k)_{k \neq i,j}$. This makes interpretation much clearer. Finally, under the GGM assumption, we have the explicit formula: $corr(X_i, X_j | (X_k)_{k \neq i,j}) = - \frac{K_{ij}}{\sqrt{K_{ii} K_{jj}}}$, where $K := \Sigma^{-1}$ is the inverse of the unknown covariance matrix. This means that the conditional correlations graph between the components of $X$ is entirely described by a single matrix parameter, $K$. Moreover the graph and $K$ have the exact same sparsity structure. With this property in mind, the author of \cite{dempster1972covariance} introduced the idea of Covariance Selection which consists of inferring - under a Gaussian assumption - a sparse estimation $\widehat{K}$ of $K$ and interpreting its sparsity structure as a conditional dependency network.\\
Subsequently, many authors have proposed their own estimators $\widehat{K}$. In \cite{meinshausen2006high}, a local edge selection approach that solves a LASSO problem on each node is introduced. It was noticeably followed by \cite{giraud2008estimation, giraud2012graph}, who developed the GGMselect algorithm, a practical implementation of this approach coupled with a model selection procedure. We call these methods "local", since they focus on solving problems independently at each node, and evaluating performances with an aggregation of nodewise metrics. Other works within the local paradigm have proposed Dantzing selectors \cite{yuan2010high}, constrained $l_1$ minimisation \cite{cai2011constrained}, scaled LASSO \cite{sun2013sparse}, or merging all linear regression into a single problem \cite{rocha2008path}. On a different note, the authors of \cite{yuan2007model} and \cite{banerjee2008model} considered a more global paradigm where the estimator is solution of a single $l_1$-penalised log-likelihood optimisation problem, that has the form of Eq.~\eqref{eq_likelihood_l1}.
\begin{equation}\label{eq_likelihood_l1}
    \widehat{K} :=\argmax{\widetilde{K} \succ 0} \mathcal{L}\parent{\widetilde{K}} - \rho \sum_{i<j} \det{\widetilde{K}_{ij}} \, .
\end{equation}
We call this point of view "global" since the likelihood estimates at once the goodness of fit of the whole proposed matrix. The introduction of problem \eqref{eq_likelihood_l1} generated tremendous interest in the GGM community, and in its wake, many authors developed their own numerical methods to compute its solution efficiently. 
A few notable examples are block coordinate descent for the Graphical Lasso algorithm (GLASSO) of \cite{friedman2008sparse}, Nesterov's Smooth gradient methods \cite{d2008first}, Interior Point Methods (IPM) \cite{li2010inexact}, Alternating Direction Methods of Multipliers (ADMM) \cite{yuan2009alternating, scheinberg2010sparse}, Newton-CG primal proximal point \cite{wang2010solving}, Newton's method with sparse approximation \cite{hsieh2011sparse}, Projected Subgradient Methods (PSM) \cite{duchi2012projected}, and multiple QP problems for the DP-GLASSO algorithm of \cite{mazumder2012graphical}. The theoretical properties of the solutions to Eq.~\eqref{eq_likelihood_l1} are studied in \cite{rothman2008sparse}, \cite{lam2009sparsistency} and in \cite{ravikumar2011high}. Other methods within the global paradigm include \cite{fan2009network}, with penalties other than $l_1$ in \eqref{eq_likelihood_l1}, and \cite{li2012sparse}, with a RKHS estimator.\\
More recent works have proposed more involved estimators, defined as modifications of already existing solutions and possessing improved statistical properties, such as asymptotic normality or better element-wise convergence. The authors of \cite{ren2015asymptotic} and \cite{jankova2017honest} adapted solutions of local regression problems including \cite{meinshausen2006high}, whereas \cite{jankova2015confidence} modified the solutions of \eqref{eq_likelihood_l1}. In \cite{jankova2018inference}, the two approaches are unified with a de-biasing method applied to both local and global estimators.\\
\\
In our applications - where the number of observations $n$ is a fixed small number, usually smaller than the number of vertices $p$ - we did not find satisfaction with the state of the art methods from either the local or the global approach. On one hand, GGMselect yields surprisingly too sparse graph, missing many of the important already known edges.
On the other hand, the only solutions from the penalised likelihood problem (1) that are a decent fit for real distribution have so many edges that the information is hidden.
To interpret a graph, one would prefer an intermediary number of edges. Additionally, the low sample size setting requires a method with non-asymptotic theoretical properties.\\
\\
In this paper, we design a composite method, combining the respective strengths of the local and global approaches, with the aim of recovering graphs with a more reasonable amount of edges, that also achieves a better quantitative fit with the data. We also prove non-asymptotic oracle bounds in expectation and probability on the solution.\\
\\
To measure the goodness of fit, many applications are interested in recovering the true graph structure and focus on the "sparsistency". In our case, the presence or absence of an edge is not sufficient information. The correlation amplitude is of equal interest.
Additionally, we need the resulting structure to make sense as a whole, that is to say: describe a co-variation dynamic as close as possible to the real one despite being a sparse approximation. This means that edgewise coefficient recovery - as assessed by the $l_2$ error $\norm{K - \widehat{K}}_F^2 = \sum_{i,j} (K_{i,j} - \widehat{K}_{i, j})^2$ for instance - which does not take into account the geometric structure of the graph as a whole is not satisfactory either. We want the distribution function described by the proposed matrix to be similar to the original distribution. The natural metric to describe proximity between distribution functions is Cross Entropy (CE) or, equivalently, the Kullback-Leibler divergence (KL). In the end, the CE between the original distribution and the proposed one - $\N{0, \widehat{K}^{-1}}$ - is our metric of choice. Other works, such as \cite{levina2008sparse} and \cite{zhou2011high}, have focused on the KL in the context of GGM as well.\\
\\
In the following, we quantify the shortcomings of the literature's local and global methods when the data is not abundant. The GGMselect graphs are very sparse, but consistently and substantially outperform the solutions of Eq.~\eqref{eq_likelihood_l1} in terms of KL, regardless of the penalisation intensity $\rho$. In the KL/sparsity space, the solutions of GGMselect occupy a spot of high performing, very sparse solutions that the problem \eqref{eq_likelihood_l1} simply does not reach. Additionally, the better performing solutions of \eqref{eq_likelihood_l1} are so dense that they are excessively difficult to read. Subsequently, we demonstrate that despite its apparent success, the GGMselect algorithm is held back by its model selection criterion which is far too conservative and interrupts the graph exploration process too early. This results in graphs that are not only difficult to interpret but also perform sub-optimally in terms of KL.\\
With those observations in mind, we design a simple nodewise exploration numerical scheme which, when initialised at the GGMselect solution, is able to extract a family of larger, better performing graphs. We couple this exploration process with a KL-based model selection criterion to identify the best candidates among this family. This algorithm is composite insofar as it combines a careful local graph construction process with a perceptive global evaluation of the encountered graphs. \\
We prove non-asymptotic guarantees on the solution of the model selection procedure. We demonstrate with experiments on synthetic data that this selection procedure satisfies our stated goals. Indeed, the selected graphs are both substantially better in terms of distribution reconstruction (KL divergence), and much closer to the original graph than any other we obtain with the state of the art methods. Then, we put our method to the test with two experiments on real medical data. First on a neurological dataset with multiple modalities of brain imaging data, where $n < p$. Then on biological measures taken from healthy nephrology test subjects, with $p < n$. In both cases, the results of our method correspond more to the common understanding of the phenomena in their respective fields.

 

\section{Covariance Selection within GGM}

\subsection{Introduction to Gaussian Graphical Models} Let $S_p^+$ and $S_p^{++}$ be respectively the spaces of positive semi-definite and positive definite matrices in $\R^{p \times p}$. We model a phenomenon as a centred multivariate normal distribution in $\R^p$: $X \sim \N{0_p, \Sigma}$. To estimate the unknown covariance matrix $\Sigma \in S_p^{++}$, we have at our disposal an iid sample $\parent{X^{(1)}, ..., X^{(n)}}$ assumed to be drawn from this distribution. We want our estimation to bring interpretation on the conditional correlations network between the components of $X$. No real network is truly sparse, yet it is natural to propose a sparse approximation. Indeed, this means recovering in priority the strongest direct connections and privileging a simpler explanation of the phenomenon, one we can hope to infer even with a small amount of data. Sparsity in the conditional correlations structure is equivalent to sparsity in the inverse covariance matrix $K :=\Sigma^{-1}$. Namely $K_{ij} = 0 \iff \text{Corr} \parent{X_i, X_j | (X_k)_{k \neq i,j}} = 0$. As a consequence, our goal is to estimate from the dataset a covariance matrix $\widehat{\Sigma} \in S_p^{++}$ with both a good fit and a sparse inverse $\widehat{K}$. We say that $\widehat{\Sigma} := \widehat{K}^{-1}$ is "inverse-sparse".\\
\\
In the following, we use the Cross Entropy to quantify the performances of a proposed matrix $\widehat{K}$. The CE, $H(p,q) = - \mathbb{E}_p\brack{\text{log} q(X)} = \int_x -p(x) ln(q(x)) \mu(\mathrm{d}x)$, is an asymmetric measure of the deviation of distribution $q$ with regards to distribution $p$. The CE differs from the KL-divergence only by the term $\crossent{p}{p}$, which is constant when the reference distribution $p$ is fixed. In GGM, the score $H(f_{\Sigma}, f_{\widehat{\Sigma}})$ represents how well the normal distribution with our proposed covariance $\widehat{\Sigma}$ is able to reproduce the true distribution $\N{0, \Sigma}$. We call this score the True CE of $\widehat{\Sigma}$. This metric represents a global paradigm where we explicitly care about the behaviour of the matrix as a whole. This is in contrast to a coefficient-wise recovery, for instance, which is a summation of local, nodewise, metrics. After removal of the additive constants, we get the simple formula \eqref{eq_gaussian_ce} for the CE between two centred multivariate normal distributions $\N{0, \Sigma_1}$ and $\N{0, \Sigma_2}$.
\begin{equation} \label{eq_gaussian_ce}
    \crossent{\Sigma_1}{\Sigma_2} := \crossent{f_{\Sigma_1}}{f_{\Sigma_2}} \equiv  \frac{1}{2} \big(tr\parent{\Sigma_1 K_2} - ln(\det{K_2})\big) \, .
\end{equation}
In the general case, the CE between a proposed distribution $f_{\theta}$ and an empirical distribution $\hat{f}_n = \frac{1}{n} \sum_{i=1}^n \mathds{1}_{x=X^{(i)}}$ defined from data is the opposite of the log-likelihood: $H(\hat{f}_n,f_{\theta}) = - \frac{1}{n} \text{log}\, p_{\theta}(X^{(1)}, ..., X^{(n)})$. In the GGM case, we denote the observed data  $\underline{X} := \parent{X^{(1)}, ..., X^{(n)}}^T \in \mathbb{R}^{n \times p}$, and set $S := \frac{1}{n} \underline{X}^T\, \underline{X} \in S_p^+$, the empirical covariance matrix. The opposite log-likelihood of any centred Gaussian $\N{0, \Sigma_2}$ satisfies: 
\begin{equation} \label{eq_gaussian_likelihood}
    \crossent{S}{\Sigma_2} := \crossent{\hat{f}_n}{f_{\Sigma_2}} \equiv  \frac{1}{2} \big(tr\parent{S K_2} - ln(\det{K_2})\big) \, ,
\end{equation}
similar to Eq.~\eqref{eq_gaussian_ce}. As a result, we adopt an unified notation. Details on calculations to obtain these formulas can be found in Section \ref{appendix:CE}.\\
\\
We use the following notations for matrix algebra, let $A$ be a square real matrix, then: $\det{A}$ denotes the determinant, $\norm{A}_* := tr\parent{\parent{A^T A}^{\frac{1}{2}}}$ the nuclear norm, $\norm{A}_F := tr\parent{\parent{A^T A}}^{\frac{1}{2}} = \parent{\sum_{i,j} A_{ij}^2}^{\frac{1}{2}}$ the Frobenius norm and $\norm{A}_2 := \underset{x}{sup} \frac{\norm{Ax}_2}{\norm{x}_2} = \lambda_{max}(A)$ the spectral norm (operator norm 2) which is also the highest eigenvalue. We recall that when $A$ is symmetrical positive, then $\norm{A}_* = tr(A)$ and $\norm{A}_F = tr(A^2)^{\frac{1}{2}}$. We also consider the scalar product $\dotprod{A, B} := tr \parent{B^TA}$ on $\R^{p \times p}$. 

\subsection{Description of the state of the art} \label{section:state_of_the_art}
After its introduction, problem \eqref{eq_likelihood_l1} became the most popular method to infer graphs from data with a GGM assumption. Reducing the whole inference process to a single loss optimisation is convenient. What is more, the optimised loss is a penalised version of the likelihood - which is an estimator of the True CE - hence the method explicitly takes into account the global performances of the solution. However, even though the $l_1$ penalty mechanically induces sparsity in the solution, it does not necessarily recover the edges that best reproduce the original distribution, especially when the data is limited. Indeed, the known "sparsitency" dynamics of the solutions of \eqref{eq_likelihood_l1}, see \cite{lam2009sparsistency}, always involve a large number of observations tending towards infinity. We demonstrate in this paper that, when the sample size is small, other methods recover consequently more efficient sparse structures, inaccessible to the $l_1$ penalised problem \eqref{eq_likelihood_l1}. \\
On the other hand, the local approach of \cite{meinshausen2006high} carefully assesses each new edge, focusing on making the most efficient choice at each step. We confirm that the latter approach yields better performance by comparing the solutions of problem \eqref{eq_likelihood_l1} and GGMselect \cite{giraud2012graph} on both synthetic and real data (Sections \ref{section:exp_synth} and \ref{section:exp_real}). However, the loss optimised in GGMselect, $Crit(\G)$, see \eqref{eq:ggmselect_loss}, is an amalgam of local nodewise regression score, with no explicit regard for the overall behaviour of the matrix:
\begin{equation} \label{eq:ggmselect_loss}
    Crit(\G) := \sum_{a=1}^p \brack{\norm{X_a - \underline{X} \brack{\widehat{\theta}_{\G}}_a}^2_2 \parent{1 + \frac{pen(d_a(\G))}{n - d_a(\G)}}} \, ,
\end{equation}
where $pen$ is a specific penalty function, $d_a(\G)$ is the degree of the node $a$ in the graph $\G$, $X_a$ are all the observed values at node $a$, such that $\underline{X} = (X_1, ..., X_p) \in \R^{n \times p} $ is the full data, and:
\begin{equation} \label{eq:ggmselect_theta}
\begin{split}
     \hat{\theta}_{\G} &:= \argmin{\theta \in \Lambda_{\G}} \norm{\underline{X}(I_p - \theta)}_F^2 \\
     &= \argmin{\theta \in \Lambda_{\G}} \sum_{a = 1}^p \norm{X_a - \underline{X} \brack{\theta}_a}_2^2 \\
     &= \brace{\argmin{\theta_a \in \Lambda_{\G}^a} \norm{X_a - \underline{X} \theta_a}_2^2 }_{a = 1}^p \,,
\end{split}
\end{equation}
where $\Lambda_{\G}$ is the set of $p \times p$ matrices $\theta$ such that $\theta_{i, j}$ is non zero if and only if the edge $(i, j)$ is in $\G$, and $\Lambda_{\G}^a$ is the set of vectors $\theta_a \in \R^p$ such that $(\theta_a)_i$ is non zero if and only if the edge $(i, a)$ is in $\G$ . Note that by convention, auto-edges $(i, i)$ are never in the graph $\G$, and, in our work, $\G$ is always undirected. The full expression of $pen$ can be found in Eq. 3 of \cite{giraud2012graph}. It depends on a dimensionless hyper-parameter called $K$ which the authors recommend to set equal to 2.5. We first tried other values without observing significant change, and decided to use the recommended value in every later experiment.\\
The expression \eqref{eq:ggmselect_theta} illustrates that each nodewise coefficients $\brack{\hat{\theta}_{\G}}_a$ in the GGMselect loss are obtained from independent optimisation problems which each involve only the local sparsity of the graph in the vicinity of the node $a$, as seen in the definition of $\Lambda_{\G}^a$. In each parallel optimisation problem $\argmin{\theta_a \in \Lambda_{\G}^a} \norm{X_a - \underline{X} \theta_a}_2^2 $, the rest of the graph is not constrained, hence is implicitly fully connected. In particular, the solutions of such problems involve an estimation of the covariance matrix between the rest of the vertices that is not inverse-sparse. This can bias the procedure towards the sparser graphs since it actually implicitly measures the performances of more connected graphs. Finally, the GGMselect model selection criterion (GGMSC) explicitly penalises the degree of each node in the graph making it so that string-like structures are preferred over hubs. Empirically, we observe that with low amounts of data, graphs with hubs are consistently dismissed by the GGMSC. Overall, we expect the selected solutions to be excessively sparse, which experiments on both synthetic and real data in Sections \ref{section:exp_synth} and \ref{section:exp_real} confirm.

\subsection{Graph constrained MLE} Even though a covariance matrix $\Sigma$ uniquely defines a graph with its inverse $K$, the reciprocal is not true. To a given graph $\G := (V, E)$, with vertex set $V$ and edge set $E$, corresponds a whole subset $\Theta_{\G}$ of $S_p^{++}$:
\begin{equation*}
    \Theta_{\G} := \brace{\widetilde{\Sigma} \in S_p^{++} \Big| \forall i \neq j, \;  (i,j) \notin E \Rightarrow \parent{\widetilde{\Sigma}^{-1}}_{ij} = 0 }\, .
\end{equation*}
When data is available, the natural matrix representing $\G$ is the constrained MLE:
\begin{equation} \label{eq_mle}
    \widehat{\Sigma}_{\G} := \argmax{\widetilde{\Sigma} \in \Theta_{\G}} p_{\widetilde{\Sigma}}(X^{(1)}, ..., X^{(n)}) =  \argmin{\widetilde{\Sigma} \in \Theta_{\G}} \crossent{S}{\widetilde{\Sigma}}\, .
\end{equation}
The existence of the MLE is not always guaranteed (see \cite{dempster1972covariance, uhler2012geometry}). When $n<p$, no MLE exists for the more connected graphs. However, in this paper, we design a procedure that can propose a MLE for any $n$ and any graph without computation errors. To tackle the issue of existence, we add a very small regularisation term to the empirical covariance matrix $S$. This leads to solving:
\begin{equation} \label{eq_def_max_vrais_lambda}
    \widehat{\Sigma}_{\G, \lambda} := \underset{\widetilde{\Sigma} \in \Theta_{\G}}{\text{argmin}} \, \crossent{S + \lambda I_p}{\widetilde{\Sigma}} \, .
\end{equation}
$\lambda$ is not a true hyper parameter of the model. Its value is set once and for all, and as small as possible as long as the machine still recognises $S + \lambda I_p$ as invertible. Typical values range between $10^{-7}$ and $10^{-4}$. This trick changes little for the already existing solutions. Indeed, if $\widehat{\Sigma}_{\G}$ solution of Eq.~\eqref{eq_mle} exists, we observe empirically that for small values of $\lambda$: $\widehat{\Sigma}_{\G} \simeq \widehat{\Sigma}_{\G, \lambda}$. On the other hand, if no solution $\widehat{\Sigma}_{\G}$ to Eq.~\eqref{eq_mle} exists, then we now are able to propose a penalised MLE $\widehat{\Sigma}_{\G, \lambda}$, thus avoiding degenerated computations. From now on, the MLE we use are always solutions of \eqref{eq_def_max_vrais_lambda}. We will omit the index $\lambda$ and keep the notation $\widehat{\Sigma}_{\G}$ for the sake of simplicity.


\subsection{Our composite algorithm} 
The exploration steps of our method are a variation of the local paradigm of \cite{meinshausen2006high}. First, we use the GGMselect solution as initialisation. Then we add edges one by one: at each step, for each vertex independently, we run a sparse linear regression using as predictors the vertices that are not among its neighbours yet, and as target the residual of the linear regression between the value on the vertex and its neighbours. With these regressions, each vertex proposes to add to the current graph an edge between them and their new best predictor. Here however, we deviate from the local paradigm by using a global criterion - the out of sample likelihood of the whole resulting new matrix - to evaluate each proposition and select one edge among these candidates. We end this exploration procedure after a fixed number of steps, the result is a family of gradually more connected graphs. The final selection step is done with a global metric: we pick, among the so constructed family, the graph minimising the Cross Validated (with fresh data) Cross Entropy. See \figurename~\ref{algo:composite} for the details.\\
In the spirit of \cite{ren2015asymptotic, jankova2017honest, jankova2015confidence, jankova2018inference}, this method is designed to complete an already existing efficient, but sparse, solution. As a result, it is sensitive to the initial graph.
\begin{figure}
    \centering
    \begin{algorithmic}
    \STATE{\textbf{Inputs:} The \textit{train} set are all the observations available for graph inference, Nb of steps T fixed in advance.}\\
    \STATE \textbf{Start:}\\
    \textcolor{OliveGreen}{$\bullet \,$}Run GGMselect on the \textit{train} set to get the initial graph $\mathcal{G}_0 = (V, E_0)$;\\
    Partition the \textit{train} set into a \textit{validation} set and \textit{exploration} set;\\
    \FOR{$t = 1,.., T$}
        \STATE{
        Partition randomly the \textit{exploration} set into a \textit{learning} set and an \textit{evaluation} set;\\
        Compute the empirical covariance  $S_{eval}^t$ from the \textit{evaluation} set;\\
        \# \textit{We then "ask" each node for its desired next neighbour:}
        
        \FOR{$a \in V$ vertex of $\mathcal{G}_{t-1}$}
            \STATE{
            \textcolor{OliveGreen}{$\bullet \,$} Let $N_{t-1}(a)$ be the set of neighbours of $a$ in $\mathcal{G}_{t-1}$ and $F_{t-1}(a) := V \setminus \brace{N_{t-1}(a) \cup \brace{a}} $ the remaining vertices;\\
            \textcolor{OliveGreen}{$\bullet \,$} Run on the \textit{learning} set the linear regression with the vector $X_a$ of the values on $a$ as the target, and the vectors $\brace{X_s | s \in  N_{t-1}(a)}$ on the neighbour nodes as predictors. Let $\Tilde{X}_a$ be the residual of this regression;\\
            \textcolor{OliveGreen}{$\bullet \,$} Run on the \textit{learning} set one step of the LARS algorithm of \cite{efron2004least}, with $\Tilde{X}_a$ as the target, and the remaining  $\brace{X_s | s \in  F_{t-1}(a)}$ as predictors. Call $c_t(a) \in F_{t-1}(a)$ the index of the feature chosen by LARS;
            }
        \ENDFOR\\
        {\#} \textit{We now have $p$ potential new edges $\brace{(a, c_t(a))}_{a \in V}$ some of which can be identical}\\
        {\#} \textit{We give priority to mutual selections: when $c_t(c_t(a)) = a$}
        \IF{$\brace{(a, c_t(a))}_{c_t(c_t(a)) = a} \neq \emptyset$}
            \STATE{
            Let $\mathcal{C} = \brace{(a, c_t(a))}_{c_t(c_t(a)) = a}$ be our set of candidate edges;\\
            {\#} \textit{We keep only the mutual selections}
            }
        \ELSE 
            \STATE{
            Let $\mathcal{C} = \brace{(a, c_t(a))}_{a \in V}$ ;\\
            {\#} \textit{No mutual selection $\Rightarrow$ keep the whole set}
            }
        \ENDIF
        \FOR{ $c\in \mathcal{C}$}
            \STATE{ 
            Compute, with the \textit{learning} set, the MLE $\widehat{\Sigma}_t^c$ from each new potential graph $\mathcal{G}_{t}^c :=  \mathcal{G}_{t-1} \cup c$;
        }
        \ENDFOR\\
        \textcolor{RedOrange}{$\bullet \,$}$c^* := \argmin{c\in \mathcal{C}} \crossent{S_{eval}^t}{\widehat{\Sigma}_t^c}$;\\ \textcolor{RedOrange}{$\bullet \,$}$\mathcal{G}_{t} := \mathcal{G}_{t}^{c^*}$;\\
       Compute, with the \textit{exploration} set, the MLE $\widehat{\Sigma}_t$ from $\mathcal{G}_{t}$;
     }
     \ENDFOR\\
     Compute, with the \textit{exploration} set, the MLE $\widehat{\Sigma}_0$ from $\mathcal{G}_{0}$;\\
     Compute the empirical covariance  $S_{val}$ from the \textit{validation} set;\\
      \textcolor{RedOrange}{$\bullet \,$}$t^* :=  \argmin{t=0,...,T} \crossent{S_{val}}{\widehat{\Sigma}_t}$;\\ 
      \textcolor{RedOrange}{$\bullet \,$} $\widehat{\mathcal{G}} := \mathcal{G}_{t^*}$;\\
    \textbf{Return:} Inferred graph $\widehat{\mathcal{G}}$. 
    \end{algorithmic}
    \caption{Composite GGM estimation. We respectively identify with green \textcolor{OliveGreen}{$\bullet \,$} or orange \textcolor{RedOrange}{$\bullet \,$} bullets the steps adhering to a {local} or {global} paradigm. {Comments} are in \textit{italics}.} 
    \label{algo:composite}
\end{figure}

\section{Oracle bounds on the model selection procedure} \label{section:th_results}
In this Section, we give non-asymptotic guarantees on the model selection step of our algorithm. We prove these results in Section \ref{appendix:proofs}. Using the statistical properties of our model selection criterion, in particular the absence of bias and convergence towards the oracle criterion, we describe the difference between the performance of the selected model and the oracle best performance ("regret"). This regret is dependent on the convergence of a Wishart random variable towards its expectation. As a result, we are able to prove non-asymptotic upper bounds in expectation and probability for the regret.
\subsection{Framework}
In this Section we define or recall the relevant concepts and notations. We recall and rephrase the definition, given in Eq.~\eqref{eq_def_max_vrais_lambda}, of the constrained Maximum Likelihood Estimator we build from a given graph $\G$:
\begin{equation*} 
\begin{split}
        \widehat{\Sigma}_{\G}(S) &= \underset{\widetilde{\Sigma} \in \Theta_{\G}}{\text{argmin}} \, \crossent{S + \lambda I_p}{\widetilde{\Sigma}} \\
        &= \underset{\widetilde{\Sigma} \in \Theta_{\G}}{\text{argmin}} \, \crossent{S}{\widetilde{\Sigma}} + \frac{\lambda}{2} \norm{\widetilde{K}}_*\, .
\end{split}
\end{equation*}
We use the Cross Validated Cross Entropy (CVCE) $\crossent{S_{val}}{\widehat{\Sigma}_{\G}(S_{expl})}$ as a criterion to pick a graph $\widehat{\G}_{CV}$ among the ones encountered. This Cross Validated criterion uses the partition of the \textit{training} set into a \textit{validation} set - used to build the estimation $S_{val}$ of the true matrix $\Sigma$ - and an \textit{exploration} set - used for the graph exploration process and to build the constrained MLE $\widehat{\Sigma}_{\G}(S_{expl})$ for each encountered graph $\G$. We compare the graph $\widehat{\G}_{CV}$ selected with CVCE with $\widehat{\G}^*$ selected with the True Cross Entropy $\crossent{\Sigma}{\widehat{\Sigma}_{\G}(S_{expl})}$ of the matrix $\widehat{\Sigma}_{\G}(S_{expl})$. We define formally those graphs: in Eq.~\eqref{eq_def_best_est_model} and \eqref{eq_model_cv}:
\begin{gather} 
    \widehat{\G}^* \in \underset{\G \in \mathcal{M}}{\text{argmin}} \brack{\crossent{\Sigma}{\widehat{\Sigma}_{\G}(S_{expl})}}\, , \label{eq_def_best_est_model}\\ 
    \widehat{\G}_{CV} \in \underset{\G \in \mathcal{M}}{\text{argmin}} \brack{ \crossent{S_{val}}{\widehat{\Sigma}_{\G}(S_{expl})}}\, , \label{eq_model_cv} 
\end{gather}
where we call $\M$ the family of graphs uncovered by the Composite algorithm.

\begin{remark}
    With the data available, the ideal model selection would be made with True Cross Entropy $\crossent{\Sigma}{\widehat{\Sigma}_{\G}(S_{train})}$ of the matrix $\widehat{\Sigma}_{\G}(S_{train})$ built from the whole \textit{train} set. Comparing ourselves to this criterion would allow to quantify the importance of having a balanced split between \textit{validation} and \textit{exploration} set. This is outside the scope of this Section. We just compare our $\crossent{S_{val}}{\widehat{\Sigma}_{\G}(S_{expl})}$ to $\crossent{\Sigma}{\widehat{\Sigma}_{\G}(S_{expl})}$. In this case, the convergence of $S_{val}$ towards $\Sigma$ is the only dynamic that matters.
\end{remark}

\subsection{Basic control} \label{section:th_basic}
In this Section, we show a general upper bound on the regret, using only the properties of the model selection criterion, and not yet the properties of the estimators. From this point on, we generally do not highlight the dependency of $\widehat{\Sigma}_{\G}$ in $S_{expl}$ to simplify notation. First of all, note that by definition we always have the lower bound on the difference of CE:
\begin{equation*}
    0 \leq \crossent{\Sigma}{\widehat{\Sigma}_{\widehat{\G}_{CV}}} - \crossent{\Sigma}{\widehat{\Sigma}_{\widehat{\G}^*}} \, .
\end{equation*}
The rest of the guarantees focus on the upper bounds for this difference.\\
From the observation that $\crossent{\Sigma}{\widehat{\Sigma}} = \crossent{S}{\widehat{\Sigma}} + \frac{1}{2} \dotprod{\Sigma - S, \widehat{K}}$, we get the control (\ref{eq_general_guarantee}) on the regret $\crossent{\Sigma}{\widehat{\Sigma}_{\widehat{\G}_{CV}}} - \crossent{\Sigma}{\widehat{\Sigma}_{\widehat{\G}^*}}$:
\begin{equation} \label{eq_general_guarantee}
    \crossent{\Sigma}{\widehat{\Sigma}_{\widehat{\G}_{CV}}} - \crossent{\Sigma}{\widehat{\Sigma}_{\widehat{\G}^*}} \leq  \frac{1}{2} \dotprod{\Sigma-S_{val}, \widehat{K}_{\widehat{\G}_{CV}} - \widehat{K}_{\widehat{\G}^*}} ,
\end{equation}
where all the MLE $\widehat{\Sigma}_{\G}$ depend only on $\G$ and $S_{expl}$. The random variable $\widehat{\G}^*$ is a function of $S_{expl}$ only, whereas $\widehat{\G}_{CV}$ depends on both $S_{val}$ and $S_{expl}$.
Since $S_{val}$ and $S_{expl}$ are independent, then:
\begin{equation*} 
    \E{ \dotprod{S_{val}, \widehat{K}_{\widehat{\G}^*}(S_{expl}) } \Big| S_{expl}} = \dotprod{\Sigma, \widehat{K}_{\widehat{\G}^*}(S_{expl}) }\, .
\end{equation*}
In the end, with $e := \E{\crossent{\Sigma}{\widehat{\Sigma}_{\widehat{\G}_{CV}}} - \crossent{\Sigma}{\widehat{\Sigma}_{\widehat{\G}^*}}}$ the expected regret, we have: 
\begin{equation} \label{eq_expectation_cv_guarantee}
    0 \leq e \leq \frac{1}{2} \E{\dotprod{\Sigma-S_{val}, \widehat{K}_{\widehat{\G}_{CV}}}}\, .
\end{equation}

\subsection{Control in expectation} \label{section:th_results_expectation}
In this Section, we use the sparsity properties of the estimator $\widehat{K}_{\widehat{\G}_{CV}}$ as well as the statistical properties of $\Sigma-S_{val}$ to obtain a more explicit control on the expected regret. In addition, we use a known concentration result to obtain an alternative control in expectation.
The result (\ref{eq_expectation_cv_guarantee}) is completely agnostic of the way the matrices $\widehat{K}_{\G} \in S_p^{++}$ are defined as long as they depend on $S_{expl}$ only. To get an order of this control, however, we use the assumption that $\widehat{\Sigma}_{\G}$ is the graph constrained MLE defined in (\ref{eq_def_max_vrais_lambda}). Let us first notice that we can ensure $\norm{\widehat{K}_{\G}}_* \leq \frac{p}{\lambda}$ thanks to our penalised definition of (\ref{eq_def_max_vrais_lambda}). Let $\Sigma_{\infty} := \max{i,j}\det{\Sigma_{ij}}$. We call $E_{\text{max}}$ the union of the maximal edge sets in $\M$, and $d_{\text{max}} = \det{E_{max}} \leq \frac{p(p-1)}{2}$ its cardinal. We underline here that, by convention, conditional correlation graphs do not contain self loops, hence the edge sets $E$ never include any of the pairs $\brace{(i,i)}_{i = 1, ..., p}$. We then get the control (\ref{eq_cv_control_order}) by using Cauchy-Schwartz's inequality in (\ref{eq_expectation_cv_guarantee}).
\begin{proposition} \textit{With the previously introduced notations, if the set $E_{max}$ is independent of the \textit{exploration} empirical matrix $S_{expl}$, we have:}
\begin{equation} \label{eq_cv_control_order}
0 \leq e \leq \frac{\Sigma_{\infty}}{\lambda \sqrt{2}} \frac{\parent{p + 2 d_{max}}^{\frac{1}{2}} p}{\sqrt{n_{val}}} \, .
\end{equation}
\textit{In the case of our Composite procedure, by construction $E_{max}$ is a random variable depending on the \textit{exploration} set. However (\ref{eq_cv_control_order}) still holds by replacing $d_{max}$ with $\E{d_{max}}$:}
\begin{equation} \label{eq_cv_control_order2}
0 \leq e \leq \frac{\Sigma_{\infty}}{\lambda \sqrt{2}} \frac{\parent{p + 2 \E{d_{max}}}^{\frac{1}{2}} p}{\sqrt{n_{val}}} \, .
\end{equation}
\end{proposition}
We can get an alternative order of the control by using known concentrations inequalities. 
\begin{proposition}
\textit{By using the Theorem 4 of \cite{koltchinskii2014concentration}, we get:}
\begin{equation}\label{eq_cv_control_lounici}
0 \leq e \leq c \frac{\lambda_{max}(\Sigma)}{\lambda} p \parent{\sqrt{\frac{p}{n_{val}}} \vee \frac{p}{n_{val}}}\, .
\end{equation}
\textit{Where $c$ is a constant independent of the problem.} 
\end{proposition}
In the end, with \eqref{eq_cv_control_order2} and \eqref{eq_cv_control_lounici}, 
we have two different upper bounds on $e$ and can use the minimum one depending on the situation.

\subsection{Control in probability} \label{section:th_results_probability}
In this Section, we use the sparsity properties of the estimator $\widehat{K}_{\widehat{\G}_{CV}}$ as well as the concentration properties of $\Sigma-S_{val}$ around 0 to obtain a control in probability (concentration inequality) on the regret.
In addition to the controls in expectation we got in (\ref{eq_expectation_cv_guarantee}) and (\ref{eq_cv_control_order}), there is in the CVCE a concentration dynamic based on the convergence rate of a Wishart random matrix towards its average. We call $\Pi_{\text{max}}$ the orthogonal projection on the set of edges $E_{\text{max}} \cup \brace{(i,i)}_{i=1}^p$. That is to say, for any matrix $M\in \R^{p \times p}, \quad \Pi_{\text{max}}(M)_{i,j} = M_{i,j}  \mathds{1}_{ (i, j) \in E_{\text{max}} \cup \brace{(i,i)}_{i=1}^p}$.  Let $W := K^{\frac{1}{2}} S_{val} K^{\frac{1}{2}}$. Then $n_{val}\, W \sim \W{n_{val}, I_p}$ is a standard Wishart random variable depending only on the \textit{validation} data, hence independent of every matrix $\widehat{K}_{\G}$. Let $P := \P{\det{\crossent{\Sigma}{\widehat{\Sigma}_{\hat{\G}_{CV}}} - \crossent{\Sigma}{\widehat{\Sigma}_{\hat{\G}^*}}} \leq \delta}$ be the probability that the regret is small. We get two different lower bounds \eqref{eq_concentration_1} and \eqref{eq_concentration_2} on $P$.
\begin{proposition} \textit{With the previously introduced notations, the two following inequalities hold:}
\begin{equation} \label{eq_concentration_1}
P \geq \P{\norm{W - I_p}_F \leq \frac{\delta}{\underset{\G}{\text{max}} \norm{\Sigma^{\frac{1}{2}} \widehat{K}_{\G} \Sigma^{\frac{1}{2}}}_F}}\, ,
\end{equation}
\begin{equation} \label{eq_concentration_2}
P \geq \P{\norm{\Pi_{\text{max}}\parent{S_{val} - \Sigma}}_F \leq \frac{\delta}{\underset{\G}{\text{max}} \norm{ \widehat{K}_{\G}}_F}}\, .
\end{equation}
\textit{Moreover, the results \eqref{eq_concentration_1} and \eqref{eq_concentration_2} hold when every probability is taken conditionally to the \textit{exploration} data or, equivalently here, conditionally to $S_{expl}$.}
\end{proposition}
If we work conditionally to the \textit{exploration} data, then $\underset{\G}{\text{max}} \norm{\Sigma^{\frac{1}{2}} \widehat{K}_{\G} \Sigma^{\frac{1}{2}}}_F$, $\underset{\G}{\text{max}} \norm{ \widehat{K}_{\G}}_F$ and $E_{max}$ are constants of the problem. In that case, the lower bound in \eqref{eq_concentration_1} only depends on the dynamic of a standard Wishart $\W{n_{val}, I_p}$. Similarly, the lower bound in \eqref{eq_concentration_2} only depends on the convergence dynamic of some coefficients of $S_{val}$ towards the corresponding ones in $\Sigma$.\\
The bound in \eqref{eq_concentration_2} has a less general formulation than \eqref{eq_concentration_1}, since the $S_{val} \mapsto \Sigma$ is a more specific dynamic than $W \mapsto I_p$. On the other hand, only the diagonal coefficients and those in $E_{\text{max}}$ need to be close, which can make a huge difference if $p$ is very large and $\M$ contains only sparse graphs and make the bound \eqref{eq_concentration_2} tighter. 

\section{Experiments on synthetic data} \label{section:exp_synth}
We show in this Section the shortcomings of the global problem \eqref{eq_likelihood_l1} of \cite{yuan2007model} and \cite{banerjee2008model} and of the local approach of \cite{meinshausen2006high} and \cite{giraud2012graph} on synthetic data. We demonstrate that - when the data is not abundant - the solutions of GGMselect consistently reproduce the true distribution much better than any solution of the global problem \eqref{eq_likelihood_l1}. In addition to being outperformed in KL divergence, the best solutions of \eqref{eq_likelihood_l1} are also very connected, consequently more than the real graph. However, we also illustrate that the solutions of GGMselect are always very sparse, regardless of the real graph. In the end, we demonstrate that our selection criterion improves both the distribution reproduction and the graph recovery of the previous two methods.
\subsection{The solutions missed by the global paradigm: a comparison of GLASSO and GGMselect
} \label{section:Glasso_GGMsel}
\begin{figure*}
    \centering
    \begin{tabular}[t]{cc}
    \centering
    \hspace{-0.5cm} 
    \begin{tabular}{ccc}
        \subfloat{
            \centering
            \includegraphics[width=0.1\textwidth]{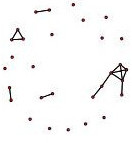}
        }
        &
        \hspace{-0.8cm} 
        \subfloat{
            \centering
            \includegraphics[width=0.1\textwidth]{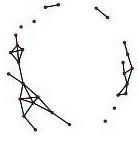}
        }
        &
        \hspace{-0.8cm} 
        \subfloat{
            \centering
            \includegraphics[width=0.1\textwidth]{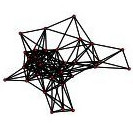} 
        }\\
    
        \subfloat{
            \centering
            \includegraphics[width=0.28\textwidth]{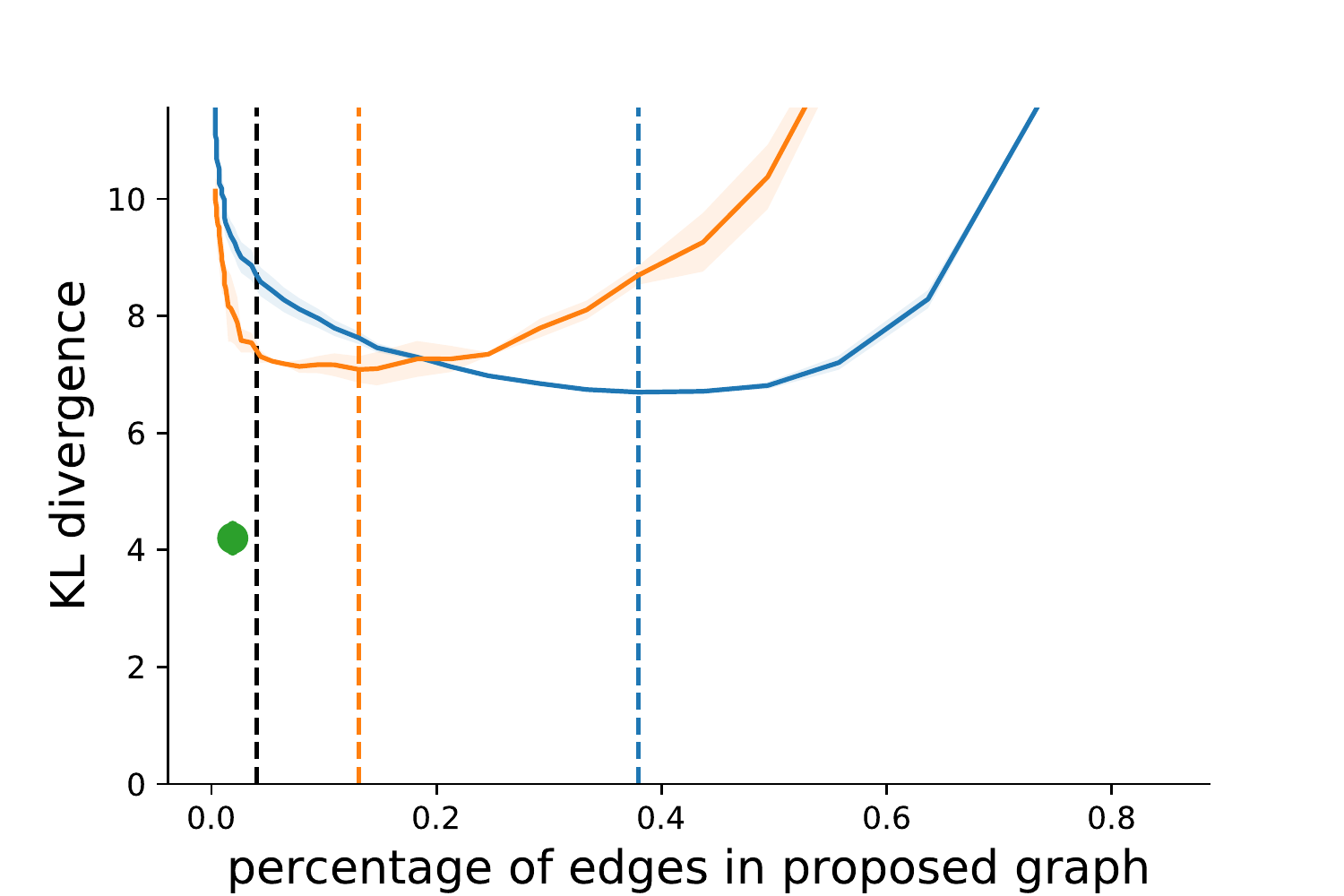}
        }
        &
        \hspace{-0.8cm} 
        \subfloat{
            \centering
            \includegraphics[width=0.28\textwidth]{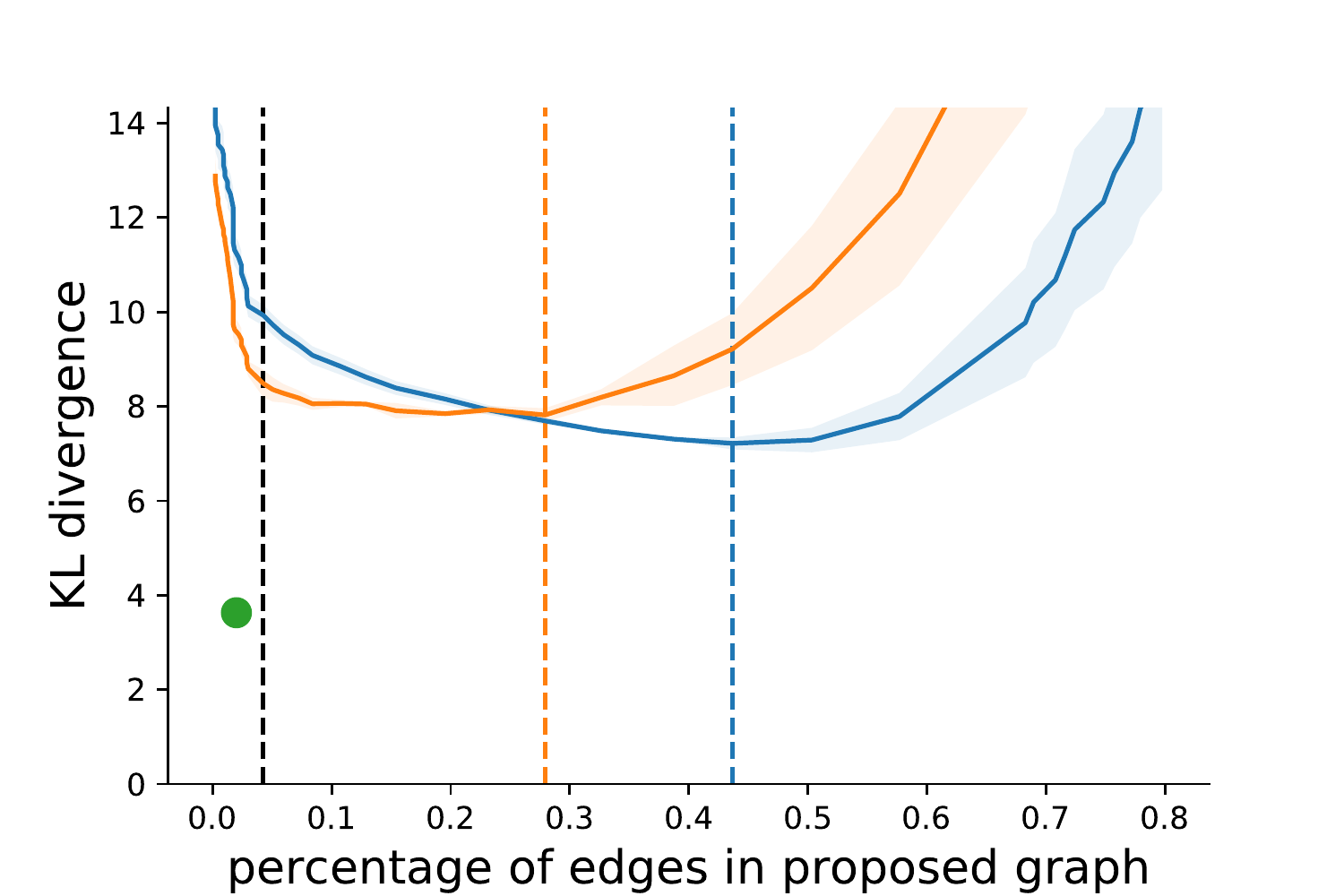}
        }
        &
        \hspace{-0.8cm} 
        \subfloat{
            \centering
            \includegraphics[width=0.28\textwidth]{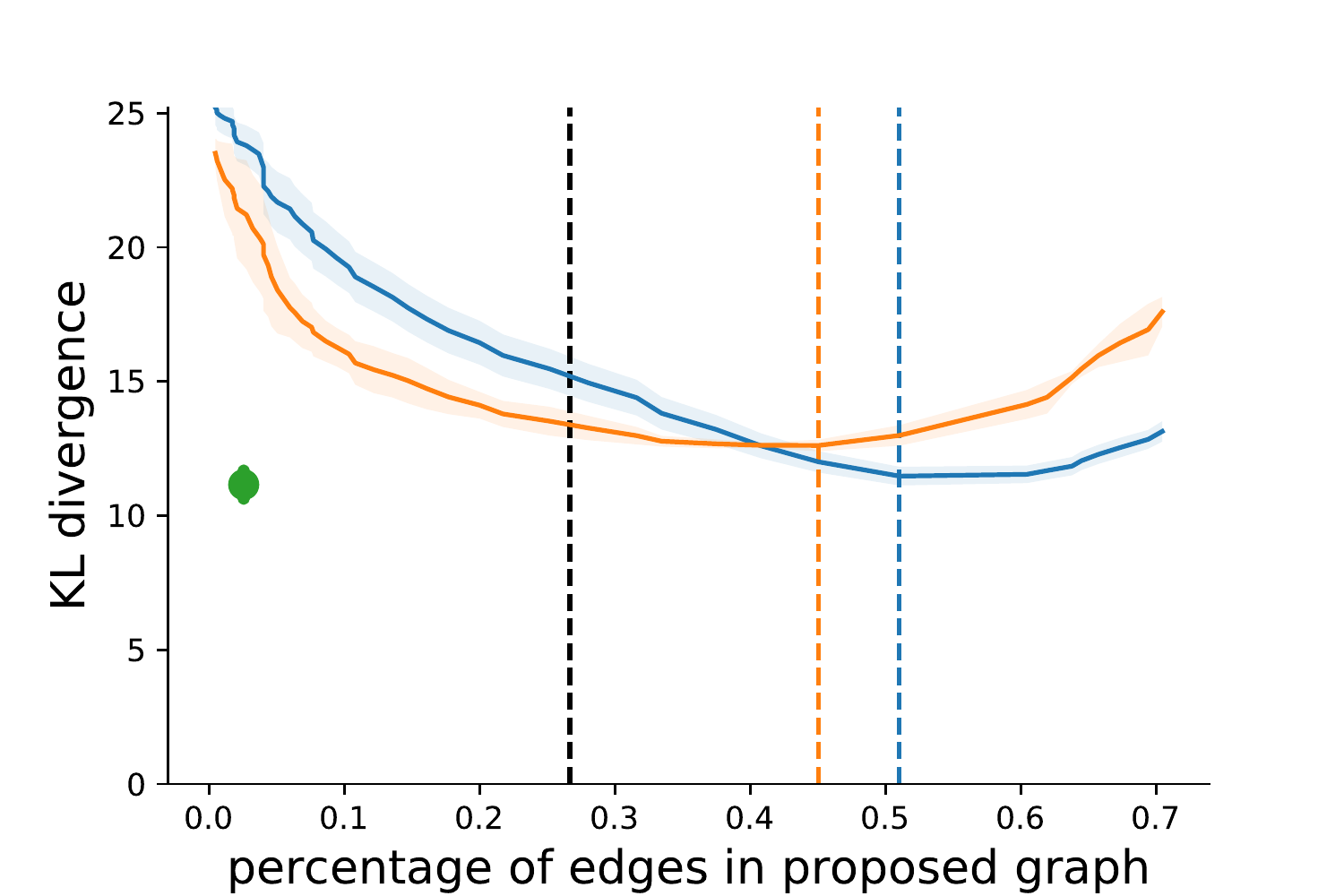} 
        }\\
        \subfloat{
            \centering
            \includegraphics[width=0.28\textwidth]{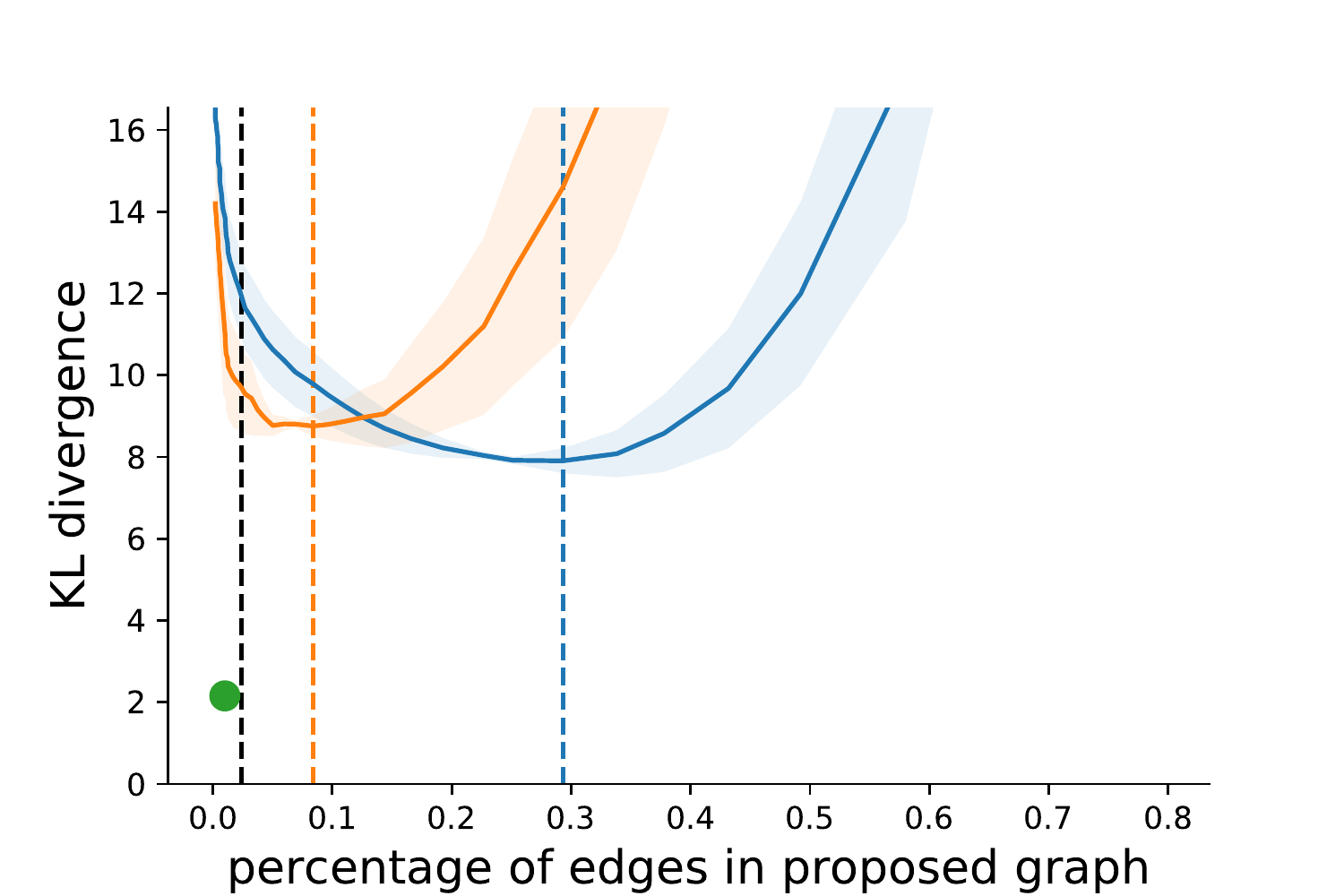} 
        }
        &
        \hspace{-0.8cm} 
        \subfloat{
            \centering
            \includegraphics[width=0.28\textwidth]{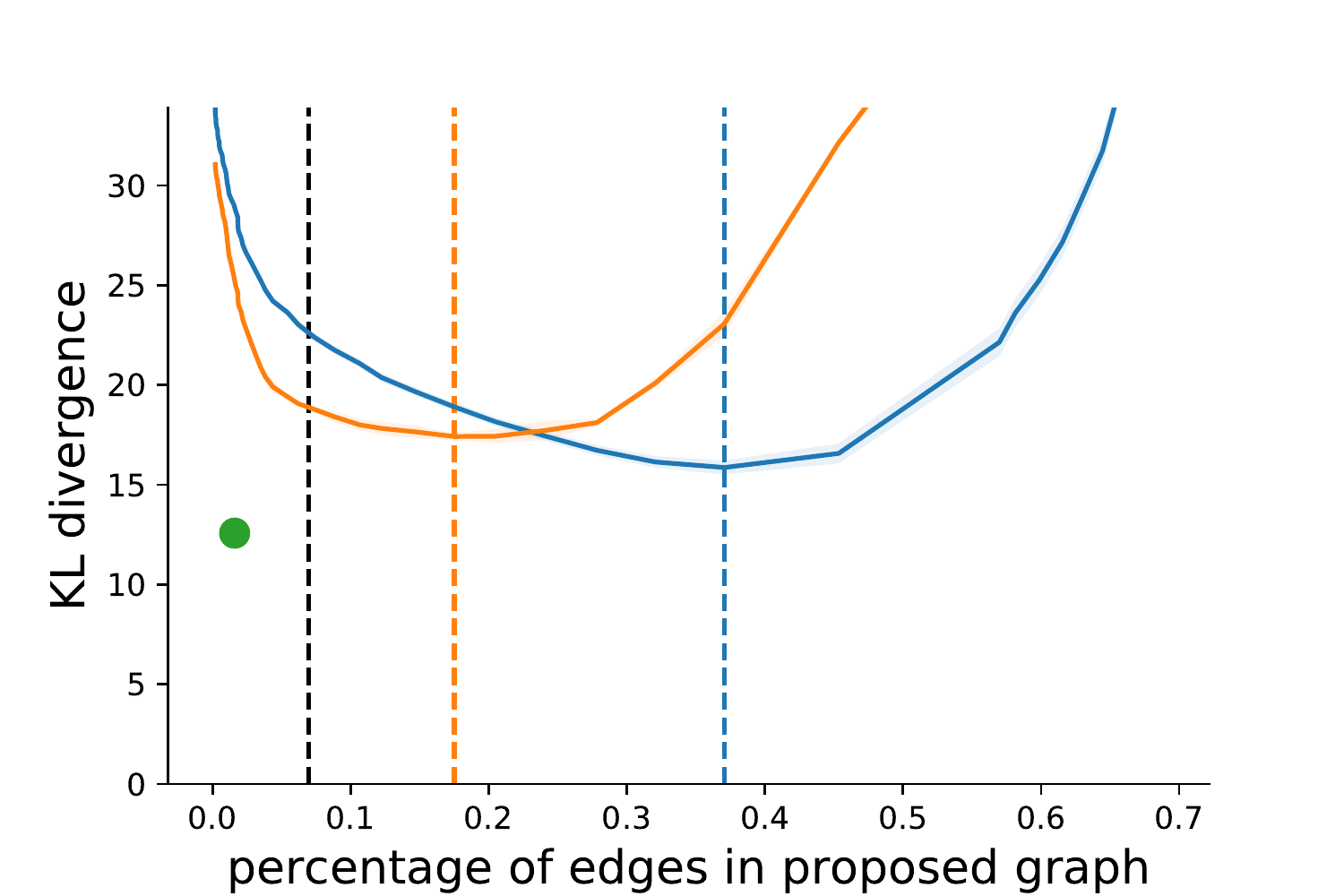} 
        }
        &
        \hspace{-0.8cm} 
        \subfloat{
            \centering
            \includegraphics[width=0.28\textwidth]{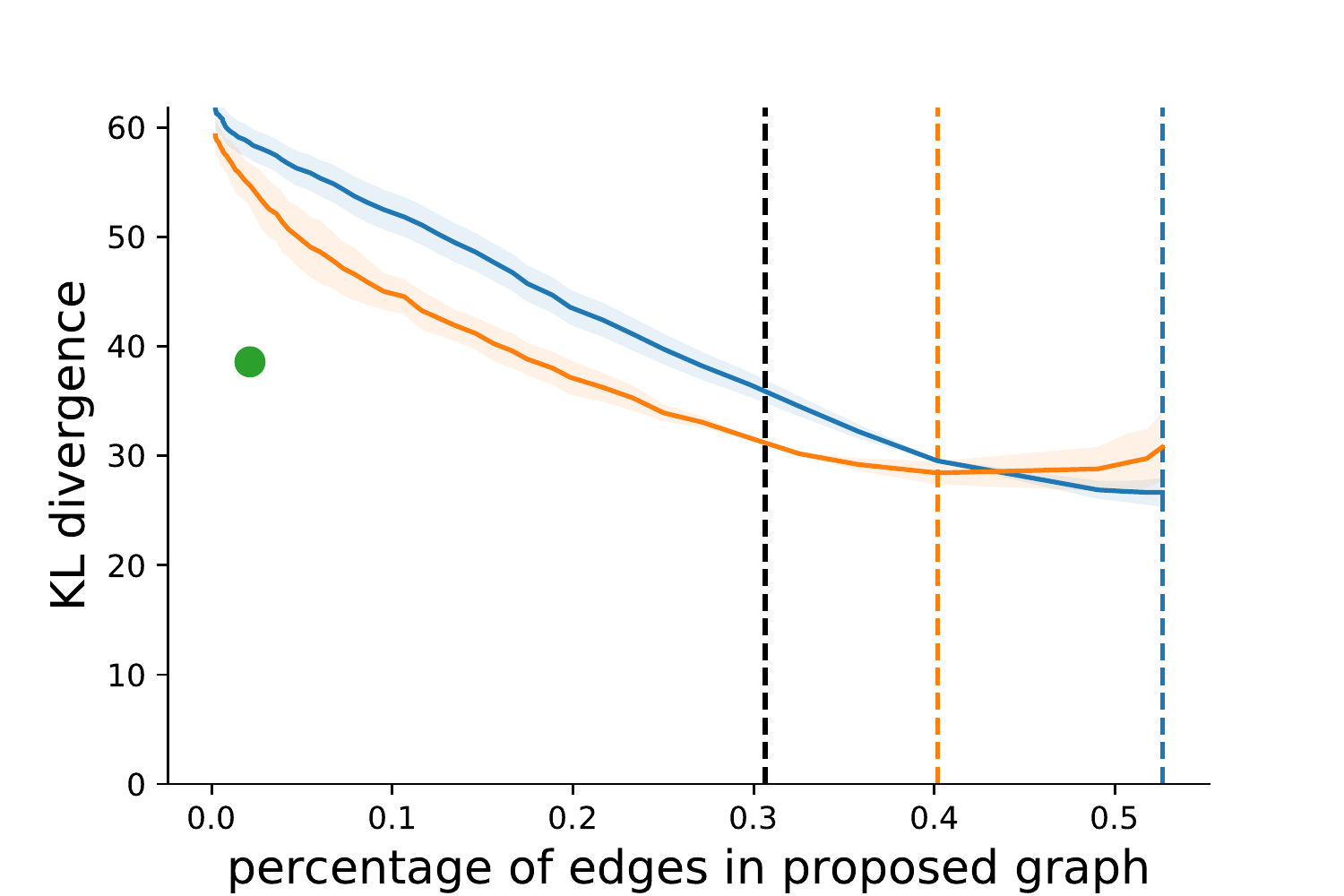} 
        }
    \end{tabular}
    &
    \hspace{-0.7cm}
    \centering
    \subfloat{
        \vspace{1cm}
        \includegraphics[width=0.18\textwidth, valign=c]{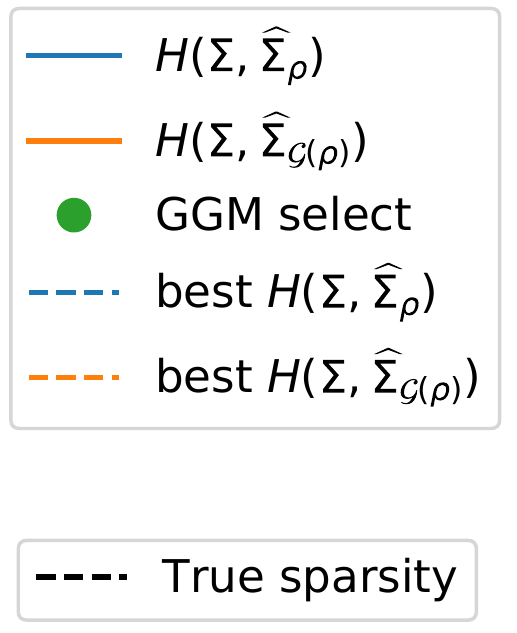} 
    }
    \end{tabular}

    \caption{Average performances as a function of the complexity for: the MLE from the GGMselect graph (green), the GLASSO solutions (blue) and the MLEs from the GLASSO graphs (orange). The average is taken over 100 simulations. In each simulation, $n=30$ data points are simulated from a given true graph, different for each subfigure. The two rows of subfigures correspond to two different graph sizes, $p = 30$ and $p=50$ vertices respectively. The three columns correspond to true graphs with different connectivity. At the top of each column, a graph illustrates the typical connectivity of the true graphs in said column.}
    \label{fig_grid_ggm_glasso}
\end{figure*}

We start by comparing the two state of the art global and local paradigms, and show that the global paradigm misses crucial solutions when the number of observations is small. We use the scikit learn, see \cite{pedregosa2011scikit}, implementation of the GLASSO of \cite{friedman2008sparse} to solve problem \eqref{eq_likelihood_l1} for any penalisation level $\rho$ and the R implementation of GGMselect, see \cite{giraud2012graph}, to represent the \cite{meinshausen2006high} approach.\\ 
We use an inverse-sparse covariance matrix $\Sigma$ fixed once and for all to generate a matrix of observations $\underline{X}$. The same observations are provided to the two methods. On \figurename~\ref{fig_grid_ggm_glasso}, we compare the True CE $H(\Sigma, \widehat{\Sigma})$ of each estimated matrix as a function of the number of non-zero, off diagonal coefficients in their inverse $\widehat{K}$ (complexity of the model). The green dot is the MLE - computed as in \eqref{eq_def_max_vrais_lambda} - under the constraints of the GGMselect graph. In the case of GLASSO, different solutions are obtained by changing the level of penalisation $\rho$ in  Eq.~\eqref{eq_likelihood_l1}. We call those solutions $\widehat{\Sigma}_{\rho}$, indexed by their penalisation intensity $\rho$. They are represented by the blue curve on \figurename~\ref{fig_grid_ggm_glasso}. All of them are inverse-sparse and define a graph we call $\G(\rho)$. The orange curve is the path of the MLEs $\widehat{\Sigma}_{\G(\rho)}$ - computed as in \eqref{eq_def_max_vrais_lambda} - refitted from those same graphs without the $l_1$ penalty of problem \eqref{eq_likelihood_l1}. They have the same inverse-sparsity as their raw solution counterparts, but do not have the extra-penalisation on the non-zero coefficients that every LASSO solution bears.\\
The three columns correspond to graphs with different connectivity - illustrated by a random example on top of each column - and the two rows have different graph sizes, $p = 30$ and $p=50$ respectively. For each simulation, the two methods were given the same $n = 30$ observations to work with, and each figure represents the average and standard deviation of 100 simulations.\\
\\
We notice that the GGMselect solution is always very sparse. When the true graph is sparse, GGMselect outperforms the penalised likelihood problem \eqref{eq_likelihood_l1} regardless of the penalty intensity. For large connected graphs, the most connected solutions of \eqref{eq_likelihood_l1} can perform better than the GGMselect solution. However GGMselect is consistently better than the equally sparse problem \eqref{eq_likelihood_l1} solution. The failure of GLASSO to reach the spot of GGMselect in the performances/complexity with any penalisation intensity - even when the MLE is refitted from the GLASSO graph without penalty - indicates that when $n$ is small, the $l_1$ penalised likelihood problem \eqref{eq_likelihood_l1} has difficulties selecting the most efficient edges. Additionally, the better performing solutions of GLASSO have many edges - usually much more than the real graph - which draws the focus away from the relevant ones and makes it difficult to get a qualitative reading of the graph.\\
When the number of observations is small, it seems that GGMselect's numerical scheme allows it to find high performing sparse graphs that problem \eqref{eq_likelihood_l1} never can. This is the type of solution we want, and the main reason why we choose to initialise our composite method from this point.

\subsection{Conservativeness of the GGMselect criterion: an example with a hub}
\begin{figure}
    \centering
        \centering
        \includegraphics[width = 0.5\textwidth]{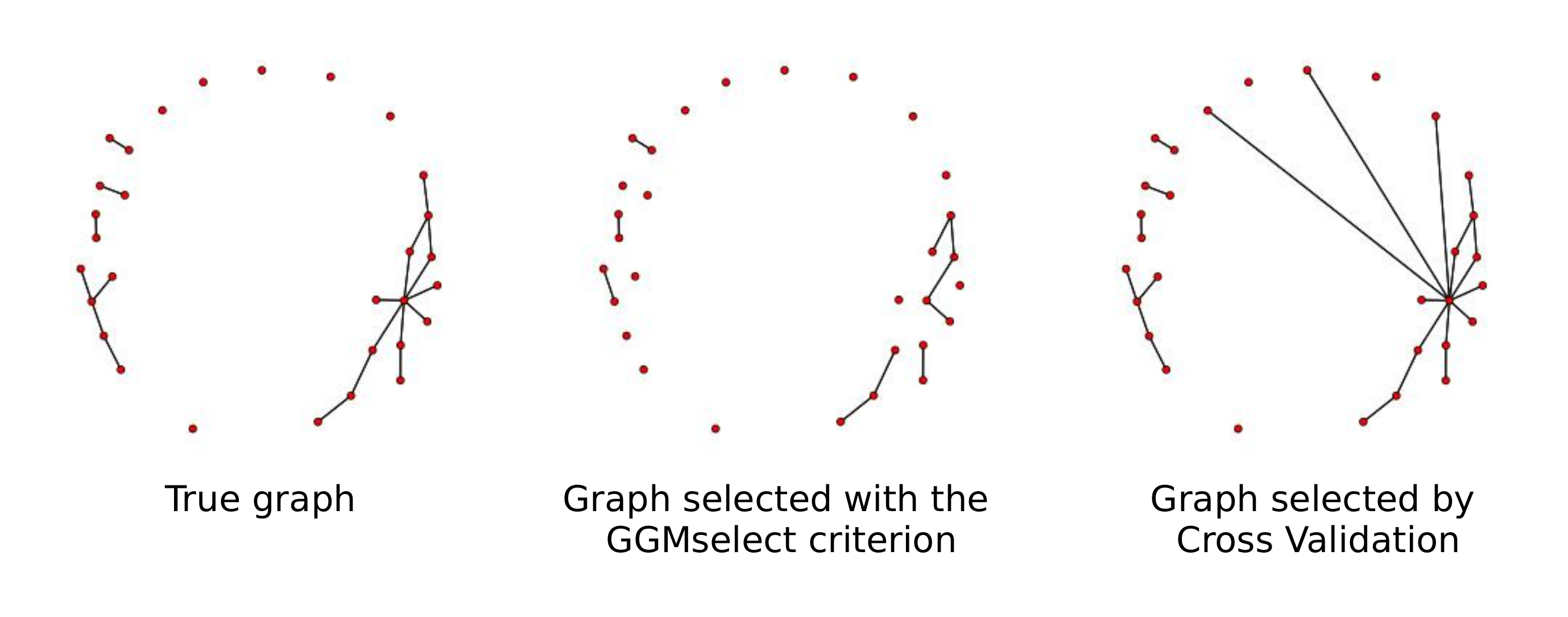}
        \caption{Graph selection in the presence of a hub. The first figure is the true graph. The second and third are the graphs respectively selected by the GGMSC and CVCE  on the same fixed graph path going from the fully sparse to the fully connected, via the GGMselect graph and the true graph} \label{fig:hub_three}
\end{figure}

We identified that GGMselect produced high quality, very sparse solutions. We argue here that they might be too sparse for their own good. \\
As discussed in Section \ref{section:state_of_the_art}, the numerical scheme of the GGMselect algorithm is based on a nodewise approach, and so is its model selection criterion. It penalises independently the degree of every node in the proposed graph. This makes it very unlikely to select graphs with a hub, i.e. a central node connected to many others. However recovering hubs is very important in conditional correlation networks. Genetic regulation networks for instance often feature hubs. With synthetic data, $n=30, p=30$, we encounter a "soft cap" effect, where it becomes very hard for GGMselect to propose a graph including a node of degree higher than 3. The penalty for such a node being too large to be compensated by the improved goodness of fit. On the other hand, we see on \figurename~\ref{fig:hub_three} that the Cross Validated Cross Entropy selects a graph which features the entire hub, and is in addition closer to the real graph regarding the remaining edges. Indeed, in the example of \figurename~\ref{fig:hub_three}, other edges than the ones forming the hub are also ignored by GGMselect. With such a behaviour of the model selection criterion when the number of observations $n$ is small, the GGMselect graphs are hard to interpret, with many key connections potentially missing.\\
Such observations motivated us to replace the GGMselect criterion with the Cross Validated Cross Entropy for graph selection. The next subsection proposes a quantitative comparison of the graphs selected by these two metrics.

\subsection{The short-sightedness of the local model selection: a comparison of the GGMselect criterion and the CVCE
}
\begin{figure}
    \centering
    \includegraphics[width = 0.5\textwidth]{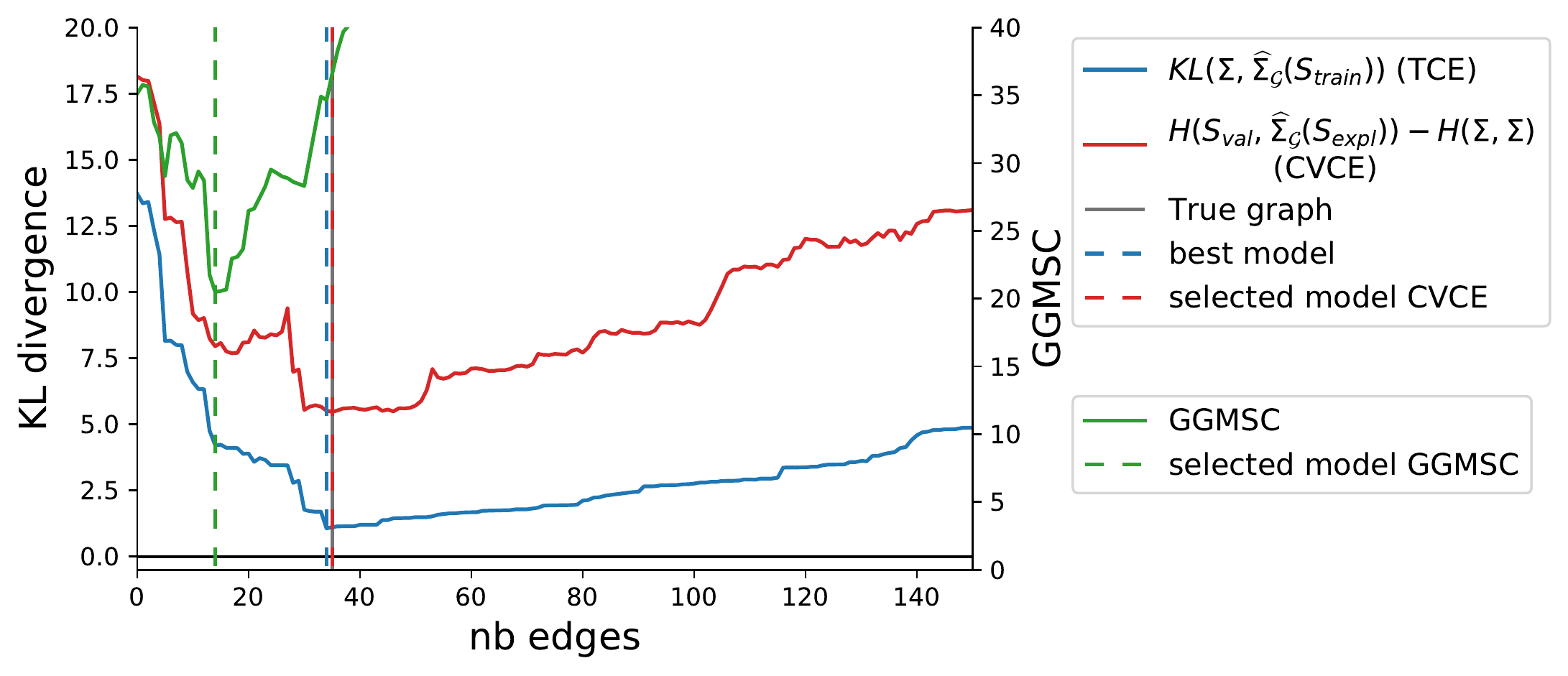}
    \caption{On a single simulation: evolution of and model selected by GGMSC (green), CVCE (red) and TCE (blue) along the fixed deterministic path. The true graph's position on that path is represented by a vertical grey line. GGMSC stops early whereas CVCE selects the true graph (the vertical grey line and the dashed red one are the same). Moreover, the CVCE graph is very close to the best graph in terms of True Cross Entropy.} \label{fig:deterministic_path_mse}
\end{figure}

\begin{figure*}
    \begin{tabular}[b]{cc}
    \centering
    \subfloat{
        \includegraphics[width=0.8\textwidth, valign=c]{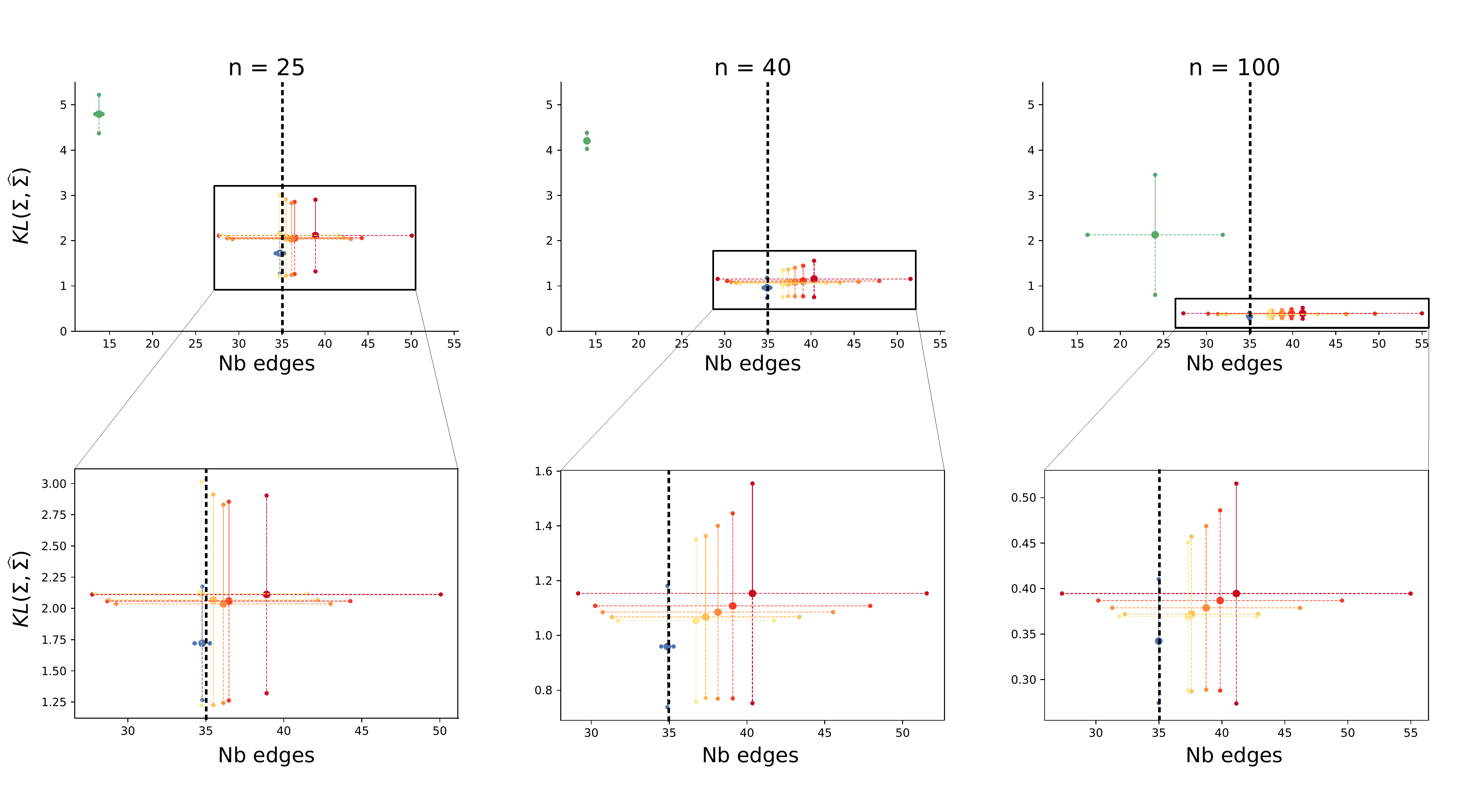}
    }
    &
    \hspace{-0.7cm}
    \centering
    \subfloat{
        \includegraphics[width=0.15\textwidth, valign=c]{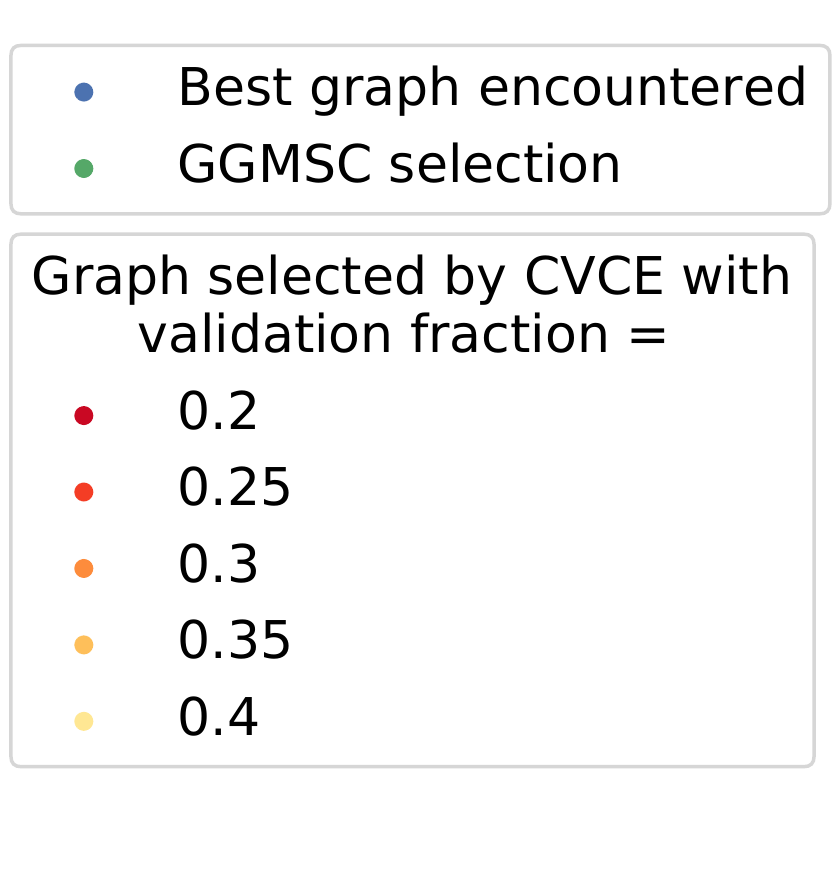}
    }
    \end{tabular}

    \caption{Average KL divergence (y axis) and complexity (x axis) of the models selected with GGMSC (green), CVCE (shades of red) and TCE (blue) on synthetic data. The sparsity level of the true graph is represented by a black dashed vertical line. The second row offers a zoomed in view of the boxed areas to focus on the CVCE and TCE models. The graphs selected by the CVCE are much closer to the best in True Cross Entropy in terms of performance and edge structure than the GGMSC one. Moreover, they are also very close to the true graph used in the simulation, even when the sample size is small.}
    \label{fig:grid_KL}
\end{figure*}

In this Section, we compare solely the model section metrics - and not the graph exploration schemes - on a fixed, shared, family of graphs. We demonstrate that our global approach to model selection yields graphs much closer to the original one and that reproduces the true distribution much better than the GGMselect criterion, which rejects the better, more connected graphs.  \\
We compare the graphs selected by our Cross Validated CE (CVCE) and the GGMSC when shown the same family of candidate graphs. We consider a given true graph ($p = 30$). We compute once and for all one GGMselect solution with $n=30$ observations drawn from this graph. With these key graphs in hand, we build manually (without the exploration scheme of \figurename~\ref{algo:composite}) a deterministic sequence of graphs. Starting from the Fully Sparse with no edges, we add one by one, and in an arbitrary order, the edges needed to reach the GGMselect graph. From there, in the same manner, we add the missing edges and remove the excess edges to reach the true graph. Finally, we add - still one by one, still in an arbitrary order - the remaining edges until the Fully Connected graph, with all possible edges. 
All the encountered graphs in this sequence constitute the fixed family of candidates to be assessed by the model selection criteria.
For each simulation, we generate $n$ observations and use them to compute the GGMSC and CVCE along the path. We make 1000 of those simulations. The GGMSC uses the full data freely, while the CVCE must split the $n$ points into the \textit{exploration} covariance $S_{expl}$, 
to compute the graph constrained MLE $\widehat{\Sigma}_{\G}(S_{expl})$, and a \textit{validation} covariance $S_{val}$ to evaluate them. This leads to different results depending on the split size. Let $S_{train}$ be the empirical covariance matrix built with the full data. We assess the performances of each graph $\G$ with the True CE (TCE) of the MLE built from $S_{train}$ under the constraints of $\G$: $H(\Sigma, \widehat{\Sigma}_{\G}(S_{train}))$. Since there is a known true $\Sigma$ we actually compute the True KL $KL(\Sigma, \widehat{\Sigma}_{\G}(S_{train}))$. This metric differs from the TCE only by a constant, hence is equivalent when ranking methods, but offers a sense of scale since the proximity to 0 in KL is meaningful. \figurename~\ref{fig:deterministic_path_mse} illustrates the behaviour on one simulation. The most noticeable trend is that the GGMSC (in green) advocates a much earlier stop than the CVCE (in red), which stops almost on the same graph as the TCE (in blue). Additionally, on that run, the graph selected by the CVCE is actually the true graph (in grey). \figurename~\ref{fig:grid_KL} represents the results over all simulations. We compare the average and standard deviation of the performances (true KL, on the y axis) and complexity (number of edges, x axis) of the models selected by the CVCE with different \textit{exploration/validation} splits (in shades of red), GGMSC (in green) and with the TCE (in blue). The three columns represent different number of available observations ($n = 25, 40, 100$) and the second row is a zoomed in view of the first. This quantitative analysis confirms that the GGMSC selects graphs that are way too sparse even when shown more complex graphs with better performances. With the performances measured in KL, relative improvement is meaningful, and we see the CVCE improving the GGMSC choice by a factor from 2 to 5, and being much closer to the oracle solution in terms of KL. Additionally, the graphs selected by CVCE are also much closer to the original one. This is especially true when a large fraction of the data ($35\%$ or $40\%$ of the \textit{training} data) is kept in the \textit{validation} set. The same results are observed with two other oracle metrics: the $l_2$ recovery of the True $\Sigma$, $\norm{\Sigma - \widehat{\Sigma}_{\G}(S_{train})}_F$, and the oracle nodewise regression $l_2$ recovery $\norm{\Sigma^{\frac{1}{2}} (I_p - \Theta_{\G}(\underline{X}_{train}))}_F $ (the oracle metric of the GGMselect authors \cite{giraud2012graph}). Those metrics also reveal that when the \textit{validation} set is small ($20 \%$), the variance of the performances of CVCE increases and it can become less reliable depending on the metric. The Figures and details on these two metrics can be found in supplementary materials.\\
This experiment illustrated how the model selection criterion of GGMselect can actually be very conservative, and even though the numerical scheme of the method explores interesting graph families, the model selection criterion might dismiss the more complex, better performing ones on them. This leads us to believe we can make substantial improvements by using the CVCE on a path built using the GGMselect solution as initialisation.

\subsection{Execution time comparison}
In this Section, we compare the runtimes of GLASSO, GGMselect and the Composite method for several values of $p$. For each $p$, 20 simulation are made, with $n = p/2$ observations each. This number of observations is an arbitrary heuristic to have both $n<p$ and $n$ increasing with $p$. \tablename~\ref{table:execution_times} synthesises the results. The runtime and complexity of the Composite method depend linearly on the number of steps chosen by the user. As seen in \figurename~\ref{algo:composite}, this number of steps is the number of graphs that are constructed and evaluated. Ideally, this sequence of graphs should be just long enough to see the Oracle (or Out of Sample) performance improve as much as they can, and stop when they start deteriorating, when the point of overfitting is reached. In this experiment, the number of steps is chosen according to an heuristic depending on the number of edges in the initialisation graph with regards to $p$. The average number of steps over the simulations is also recorded in \tablename~\ref{table:execution_times}.\\
The Composite method and GGMselect both include a model selection step, however GLASSO just returns one solution of Eq.~\eqref{eq_likelihood_l1} for one given value of the penalty parameter $\rho$. As a result, all three methods are  not strictly comparable. This was corrected in this experiment: for every simulation, the GLASSO is run on a grid of $\rho$ with as many values as the number of estimated graphs by the Composite method. We call this the "grid GLASSO".\\  
\\
\tablename~\ref{table:execution_times} shows that GGMselect is faster than the other two methods by 1 and 2 orders of magnitude in average. The Composite method is faster than the grid of GLASSOs when the dimension is small, but suffers when the dimension goes above $p=100$. The Composite algorithm has indeed a high complexity in $p$, it runs $p \times n_{steps}$ ordinary linear regression with $p-2$ features and computes then evaluates $(p + 1) \times n_{steps}$ graph constrained MLE of size $p \times p$ each.\\
\\
The algorithmic of GGMselect and GLASSO were very well optimised by their respective authors. This shows in the very fast GGMselect computations, making it a very efficient initialisation for our Composite method. However, the implementation of the Composite, see \figurename~\ref{algo:composite}, is naive and sequential. By running the linear regressions and LARS in parallel, and not re-calculating the MLE for the same graph several times, the performance would be greatly improved and closer to GLASSO.

\begin{table}[!t]
\renewcommand{\arraystretch}{1.3}
\caption{Average and (standard deviation) of the execution times of different GGM methods. The grid GLASSO compute solutions for as many values of the penalty parameter $\rho$ as there are estimated graphs (steps) in the Composite method. The last column presents the average of this number of steps/number of estimated graphs. The number of observations is $n=p/2$.}
\label{table:execution_times}
\centering
\begin{tabular}{lccccc}
p   & GGMsel (fast) & grid GLASSO & Composite & nb steps \\
\hline
30  & 0.19 (0.07) & 14.9 (8.60)  & 3.09  (1.80) & 8.4   \\
50  & 0.39 (0.03) & 62.1 (32.9)  & 16.6  (8.20) &  14.9 \\
100 & 1.66 (0.66) & 247 (135)    & 226  (138) & 26.3   \\
300 & 25.8 (1.04) & 1470 (775)   & 6847 (1453) & 40 
\end{tabular}
\end{table}

\section{Experiments on real data with the Composite GGM estimation algorithm} \label{section:exp_real}
In this Section, we present two experiments with our composite method on real data. First, we demonstrate on brain imaging data from a cohort of Alzheimer's Disease patients that it recovers the known structures better than the classical local and global methods, while also having a better Out of Sample goodness of fit with the data. Then, we showcase how it is able to describe known dynamics between factors involved in Adrenal steroid synthesis on a database of Nephrology test subjects.

\subsection{Experiment on Alzheimer's Disease patients} \label{section:Alzheimer}
We first confirm our previous observations and demonstrate the performances of the complete numerical scheme of our composite procedure on real medical data from the Alzheimer's Disease Neuroimaging Initiative (ADNI) database. We have $p=343$ features, $n = 92$ different patients. The first 240 features are measures of atrophy (MRI) and glucose consumption (PET) in the 120 areas of the cortex defined by the AAL2 map. The next 98 are two descriptors of the diffusion, fractional anisotropy and mean diffusivity, followed in the 49 regions of the JHU ICBM-DTI-81 white matter atlas. The rest of the features are basic descriptions of the patient. 

\subsubsection{Experiment} First we need a new evaluation metric. Indeed, with real data, we do not know the real covariance matrix. So we cannot anymore compute the True Cross Entropy to evaluate the inferred matrices. To replace the TCE, we keep $n = 18$ patients aside as a \textit{test} set to define a test empirical covariance matrix $S_{test}$, whereas the $n=74$ patients left constitute the \textit{train} set, used to define $S_{train}$. We evaluate an inverse-sparse covariance matrix built from $S_{train}$ with the negative Out of Sample Likelihood (OSL): $H(S_{test}, \widehat{\Sigma}_{\G}(S_{train}))$. The OSL is less absolute than the True CE, but still quantifies with no bias the goodness of fit for real data. Additionally, we cannot use a KL divergence for scale reference anymore, see Section \ref{appendix:CE} for more details.\\
The experiment run on the ADNI database is very simple: we compute the GGMselect solution and build our Composite GGM estimation procedure from it. To be fair, we also evaluate every graph our procedure encounters with the GGMSC, giving GGMselect a chance to change its mind if one of the new graphs were to fit its criterion better. In addition, we used the GLASSO algorithm of \cite{friedman2008sparse} to get the solutions of \eqref{eq_likelihood_l1} for different penalty intensity.

\subsubsection{Comparison of GLASSO and GGMselect} 
\begin{figure}
    \centering
        \includegraphics[width = 0.45\textwidth]{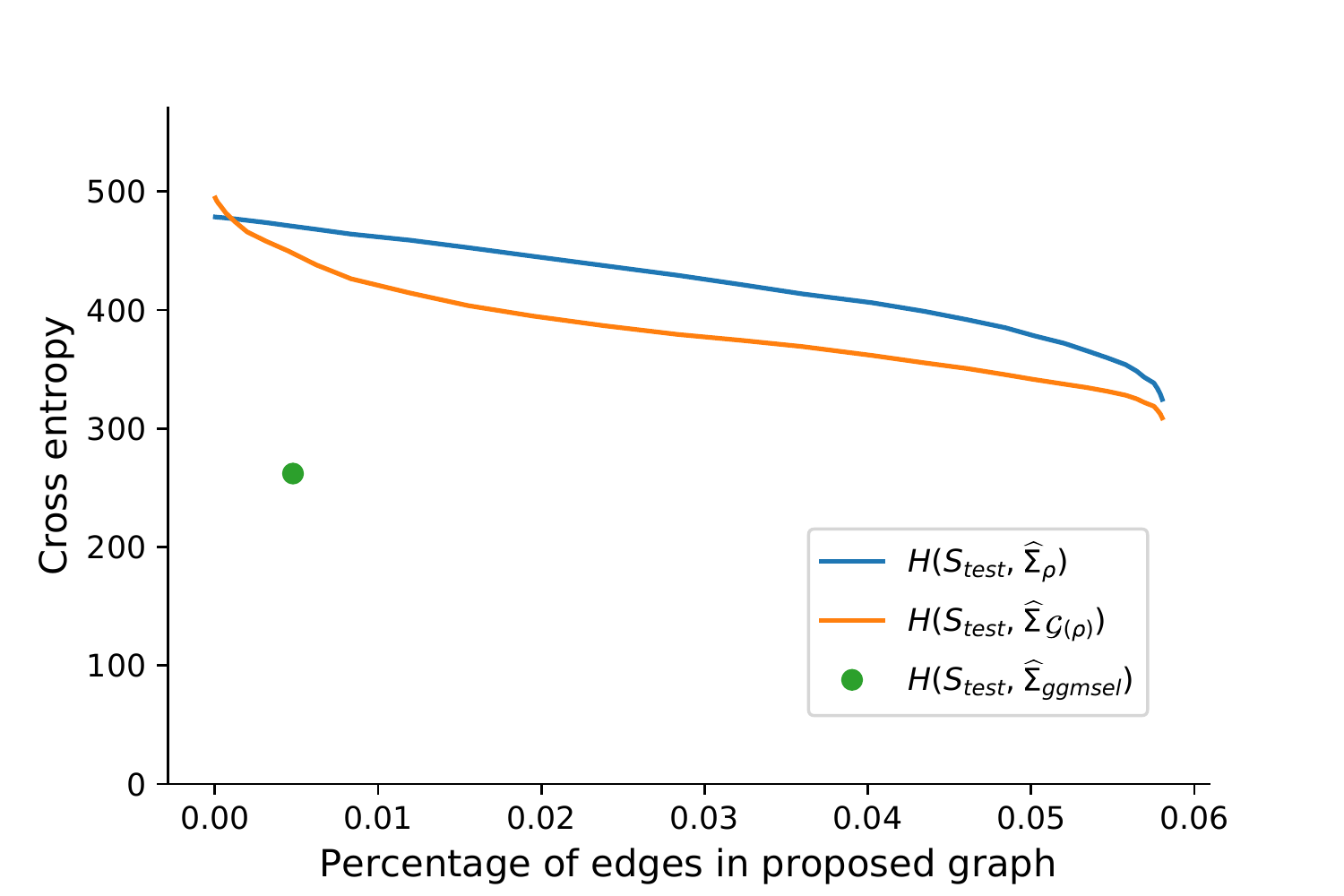}
        \caption{Out of sample performances as a function of the complexity for: the MLE from the GGMselect graph (green), the GLASSO solutions (blue) and the MLEs refitted from the GLASSO graphs (orange).} \label{fig:real_glasso_ggmsel}
\end{figure}

We confirm the observations and conclusions of Section \ref{section:Glasso_GGMsel}. \figurename~\ref{fig:real_glasso_ggmsel} shows that, even with varying penalty intensity, GLASSO does not encounter any solution with an OSL as good as GGMselect. This indicates that the optimisation problem \eqref{eq_likelihood_l1} cannot find high-performing sparse graphs in this concrete setting either. The path of GLASSO is interrupted before its completion as we have computational error with the scikit learn package at low penalty levels. We encounter such errors eventually no matter how we regularise and precondition the empirical covariance $S$. This means we do not get to see the more connected solutions of the GLASSO. This is not a problem since we already go far enough in the GLASSO path to reach unacceptably complex graphs: $6\%$ of the $\sim$ 59000 possible edges, i.e. 3500 edges for a graph with 343 nodes. By stopping early, we only consider the reasonable solutions of the GLASSO. In that case, GGMselect has a clear advantage, proposing a solution with a better Out of Sample fit with the data and only 281 edges.

\subsubsection{Comparison of GGMselect and the Composite GGM estimation algorithm} 
\begin{figure*}
    \begin{tabular}[b]{cc}
    \centering
    \hspace{-0.65cm} 
    \subfloat{
        \includegraphics[width=0.85\textwidth, valign=c]{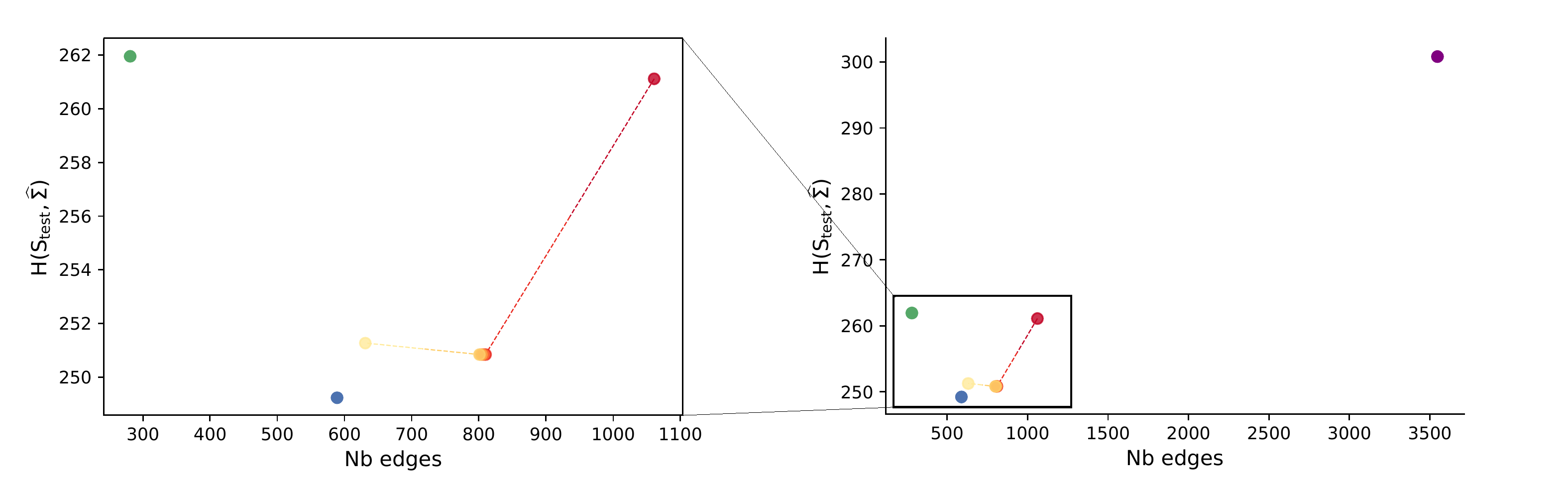}
    }
    &
    \hspace{-1.2cm}
    \centering
    \subfloat{
        \includegraphics[width=0.18\textwidth, valign=c]{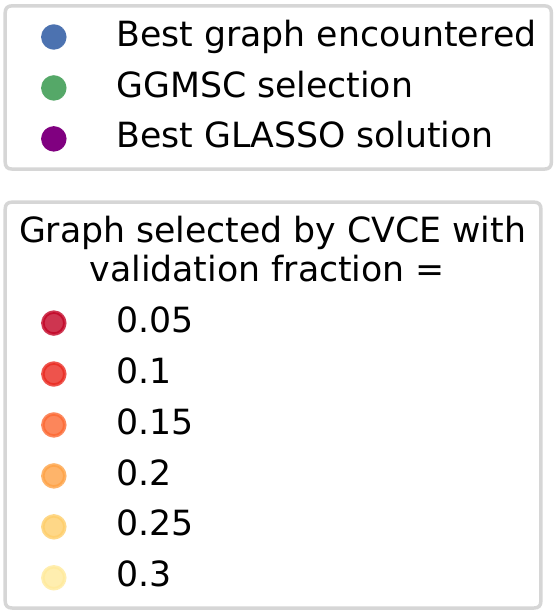}
    }
    \end{tabular}
    \caption{Out of Sample Likelihood (y axis) and complexity (x axis) of models selected by GGMSC (green), CVCE (shades of red) and OSL (blue) on real data. The right picture offers a zoomed out view to include the model selected by  OSL on the GLASSO path (purple). The left figure corresponds to the boxed area of the right figure.} \label{fig:real_KL_sparsity_glasso}
\end{figure*}

\begin{figure}
    \centering
    \subfloat{
        \includegraphics[width = 0.42\textwidth]{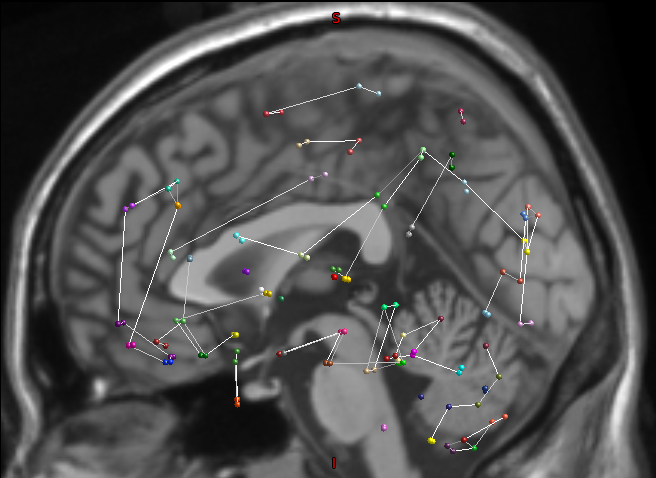}
    }\\
    \subfloat{
        \includegraphics[width = 0.42\textwidth]{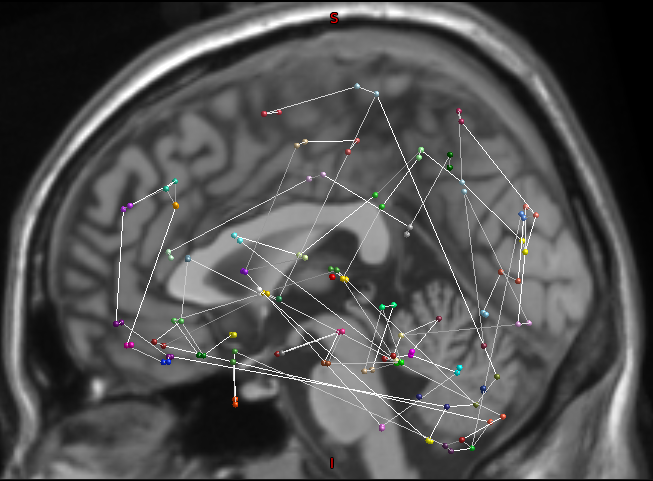}
    }\\
    \subfloat{
        \includegraphics[width = 0.42\textwidth]{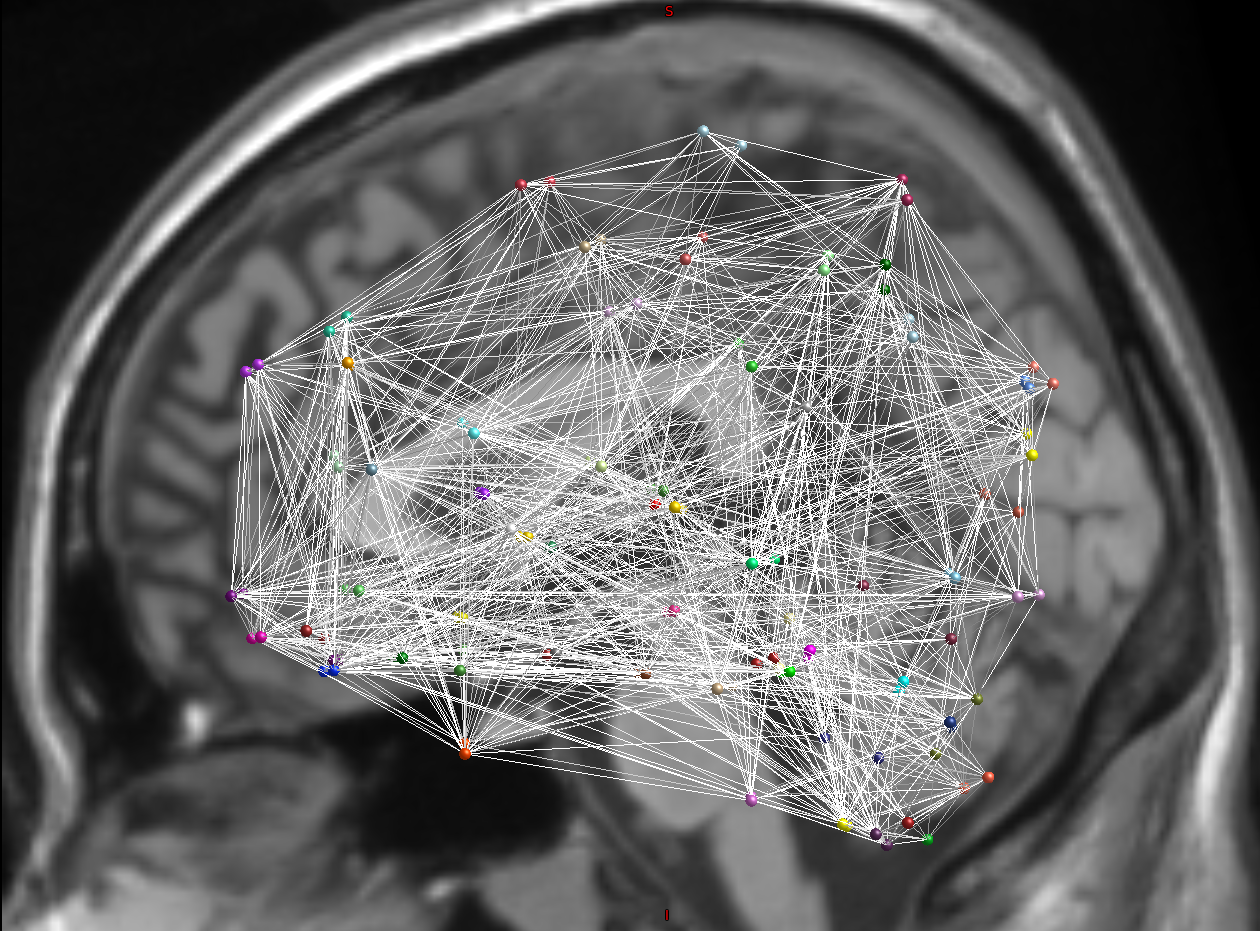}
    }
 \caption{Selected edges by GGMselect (up), our Composite method (mid) and the best Out of Sample GLASSO (down) in-between MRI measures. The perspective of the sagittal view hides the many edges between symmetrical regions. GLASSO proposes too many to allow for interpretation.} \label{fig:cortex}
\end{figure}

We represent the selected graphs on left panel of \figurename~\ref{fig:real_KL_sparsity_glasso}, with the same conventions as \figurename~\ref{fig:grid_KL}. Once again the GGMSC (green) selects a sparse model, with 281 edges over the $\sim$ 60k possible. All the reasonable \textit{validation} fractions (from 10\% to 30\%) of the CVCE (shades of red) select one out of two graphs, with both better OSL than the GGMSC one and closer to the OSL-optimum on the path (blue). Those two graphs have 589 or 813 edges respectively. This indicates that many conditional correlations were potentially missed by GGMSC, and that the CVCE graphs may propose a more complete interpretation.\\
\\
For a full comparison of the thee methods, the right panel of \figurename~\ref{fig:real_KL_sparsity_glasso} is a zoomed out view that also includes the best model obtainable with problem \eqref{eq_likelihood_l1} in terms of OSL (purple point). As we have seen, it is a very complex model with many edges. We visualise the successive improvements in Out of Sample Likelihood made first by GGMselect, with a sparser solution, then with our Composite GGM estimation procedure, with a more complete model. This experiment demonstrates the quantitative benefits of running the Composite algorithm in a High Dimension Low Sample Size setting. \\
\\
In addition to those \textit{quantitative} improvements, our method allows for a better \textit{qualitative} interpretation of the disease. \figurename~\ref{fig:cortex} represents, using the Colin 27 brain image of \cite{holmes1998enhancement} and the MRView software of \cite{tournier2012mrtrix}, the graphs selected by GGMSC and CVCE (589 edges version), as well as the best GLASSO graph in OSL ($\sim 3500$ edges). We recall that each of the methods estimates a large graph with $p=343$ vertices, a mix of different modalities measured in different areas of the cortex. The full graph cannot be displayed on an image of the cortex. For the sake of clarity, we only represent sub-parts of this one graph. On \figurename~\ref{fig:cortex}, only edges in-between the 120 MRI measures are represented. Additional views of the cortex can be found in supplementary materials. The GGMselect network is mostly composed of inter-hemispheric connections between symmetrical areas (hidden by the perspective in \figurename~\ref{fig:cortex}, see the supplementary materials for different views). These mainly reflect the symmetry of the atrophy pattern and are less informative for understanding disease process. The intra-hemispheric connections have a better interpretation potential to explain the pathology. Our algorithm reveals many more of these correlations - for instance in parietal areas, which are thought to be key hubs in the disease process - promising a more interesting description of the pathology. The GLASSO solution on the other hand, proposes many edges, making even this simple sub-graph unreadable. Similar observations can be made for connections in-between PET measures (see supplementary materials).\\
Additionally, \figurename~\ref{fig:cortex_2} shows that the GGMselect graph features absolutely no edge between MRI and PET measures, effectively proposing a model in which there is no correlation whatsoever between anatomical and functional variables, a very unlikely and unsatisfactory description. Our method on the contrary recovers a reasonable amount of edges between those two modalities. GLASSO recovers a similar number of edges in this sub-part of the graph. However, \figurename~\ref{fig:cortex} shows that it does so while having an extremely large number of edges in other regions of the graphs. Sparser GLASSO solution on the other hand, behave similarly to GGMselect and recover no edge linking MRI and PET measures, see supplementary materials. Of all these solutions, the Composite method proposes the most balanced.\\
These results suggest that our approach could be an interesting tool to study inter-regional and inter-modality dependencies in Alzheimer's Disease. This would need to be confirmed with larger populations of patients and more extensive experiments, which is out of the scope of the present paper and is left for future work.

\begin{figure}
    \centering    
    \subfloat{
        \includegraphics[width = 0.41\textwidth]{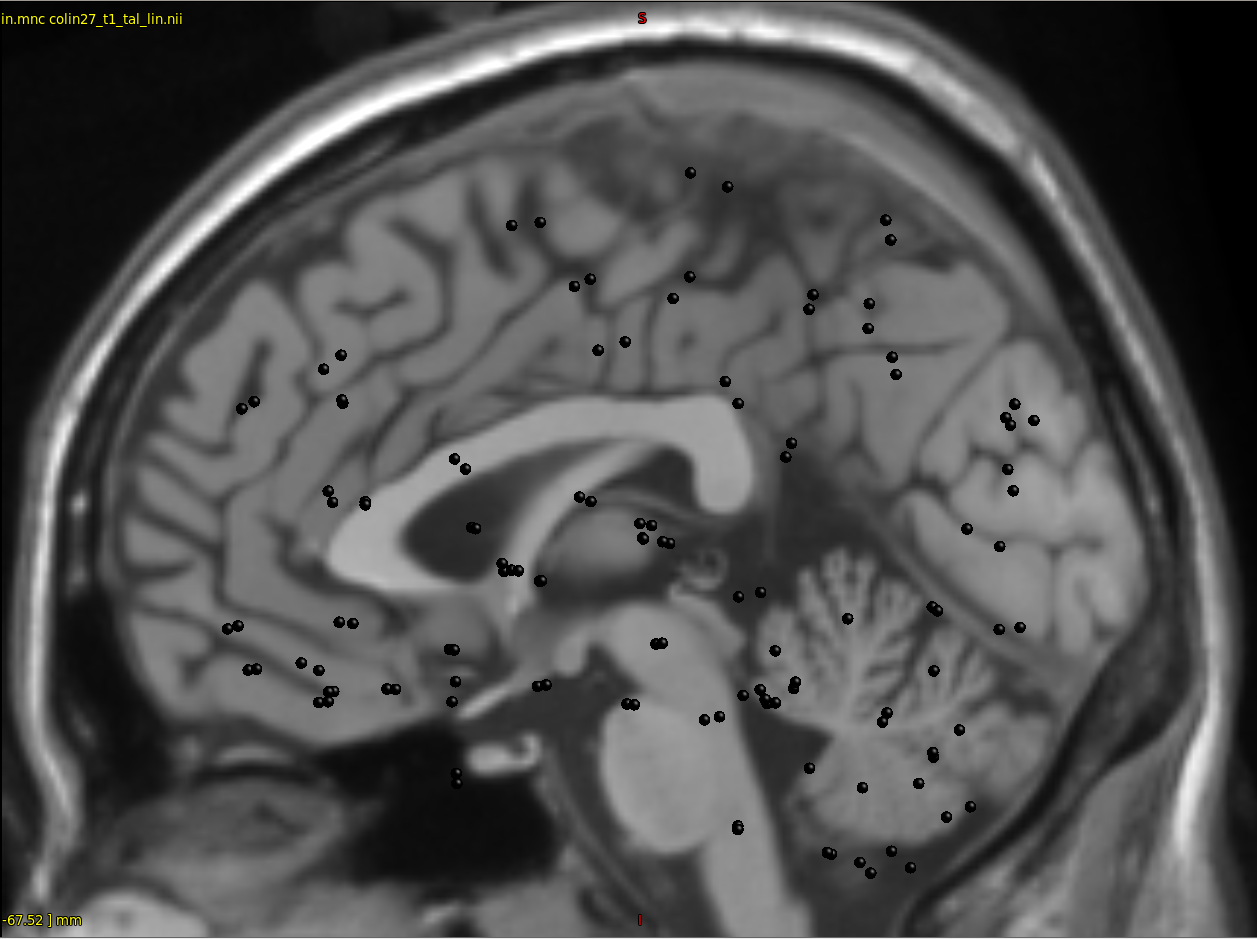}
    }\\
    \subfloat{
        \includegraphics[width = 0.41\textwidth]{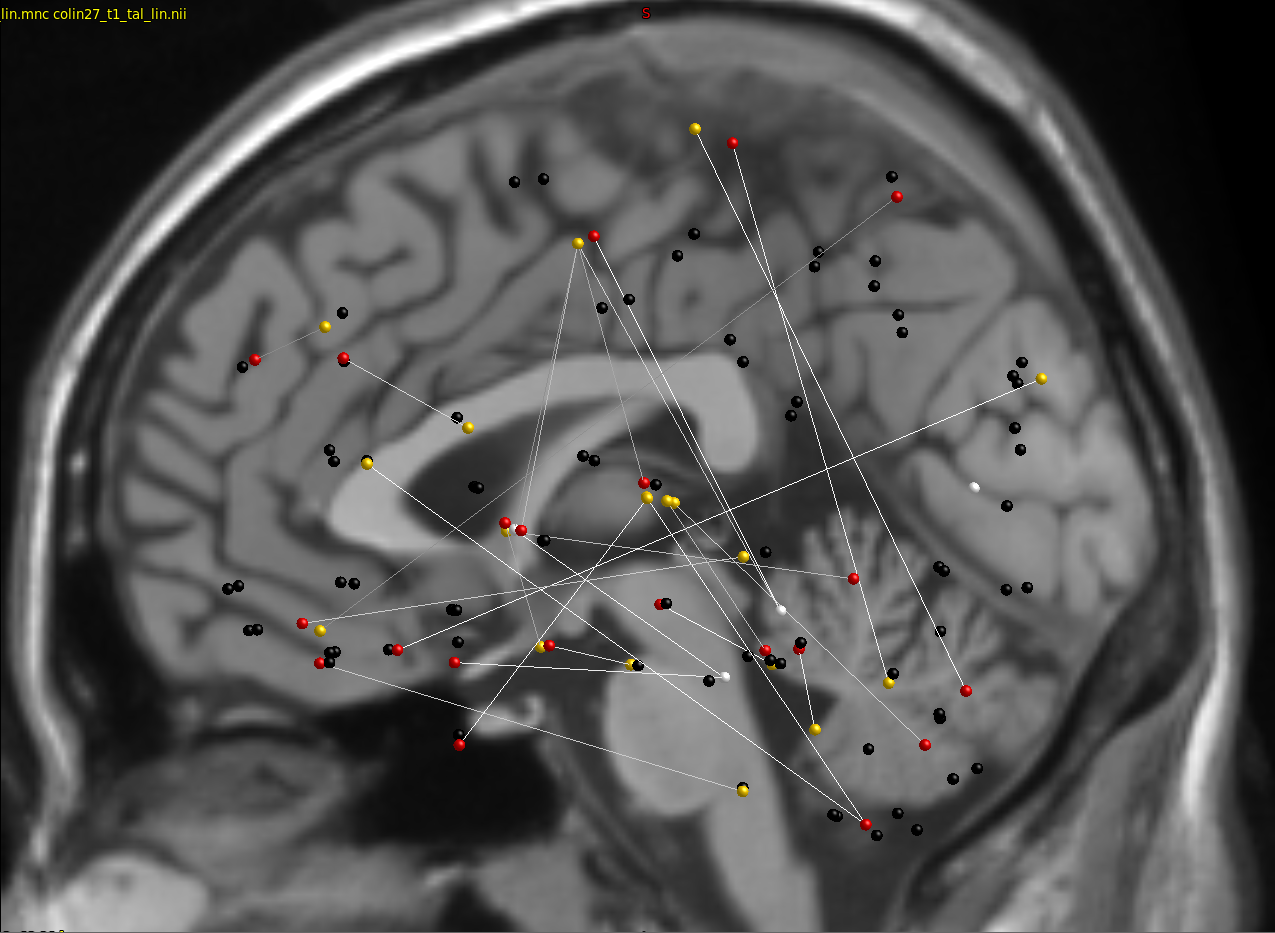}
    }\\
    \subfloat{
        \includegraphics[width = 0.41\textwidth]{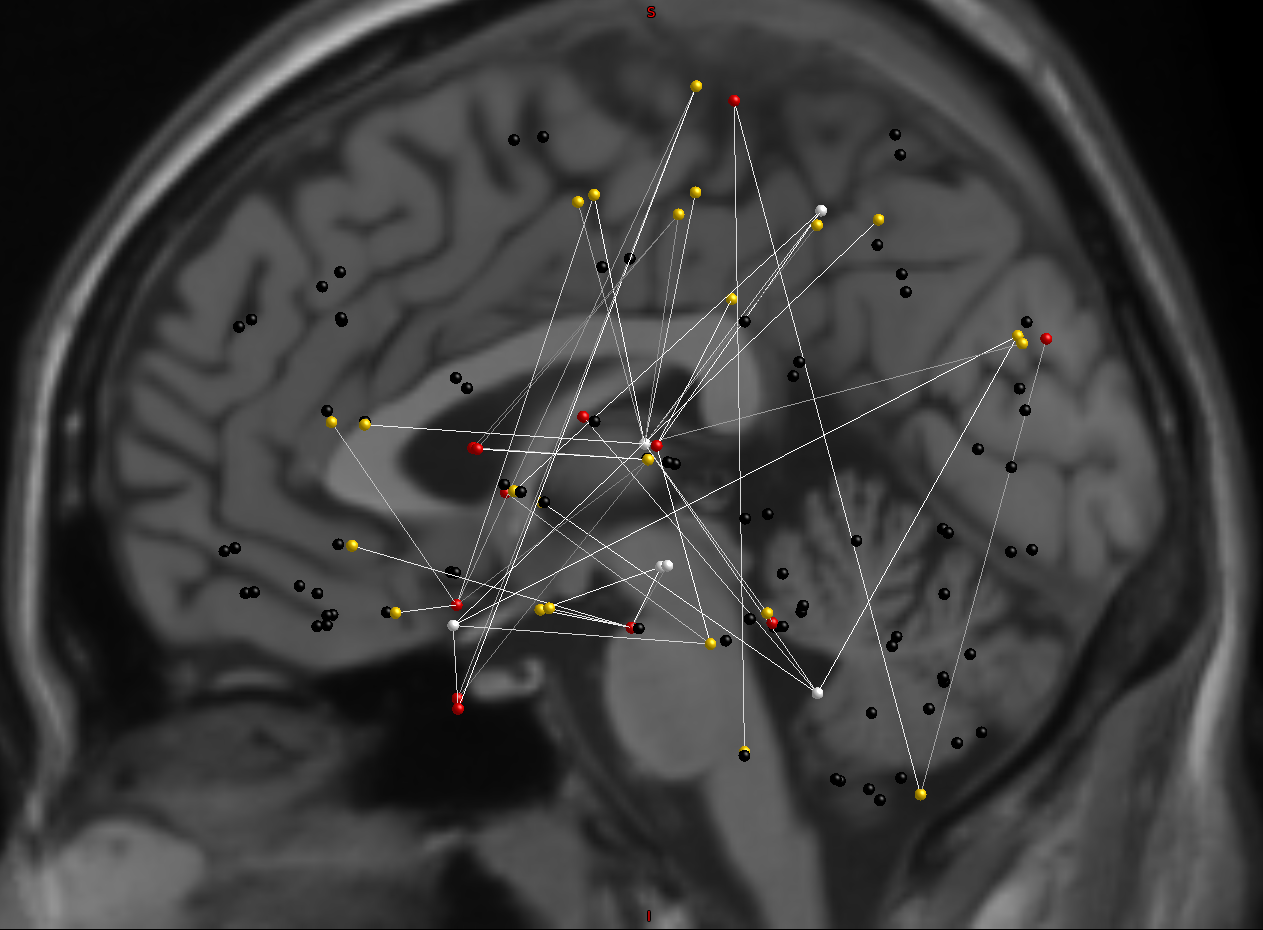}
    }
 \caption{Selected edges by GGMselect (up), our Composite method (mid) and the best Out of Sample GLASSO (down) between PET (yellow) and MRI (red) measures. GGMselect finds no connection in this sub-part of the graph, although one may expect some.} \label{fig:cortex_2}
\end{figure}

\subsection{Experiments on neprhology patients}
\begin{figure*}
    \centering
    \includegraphics[width = \textwidth]{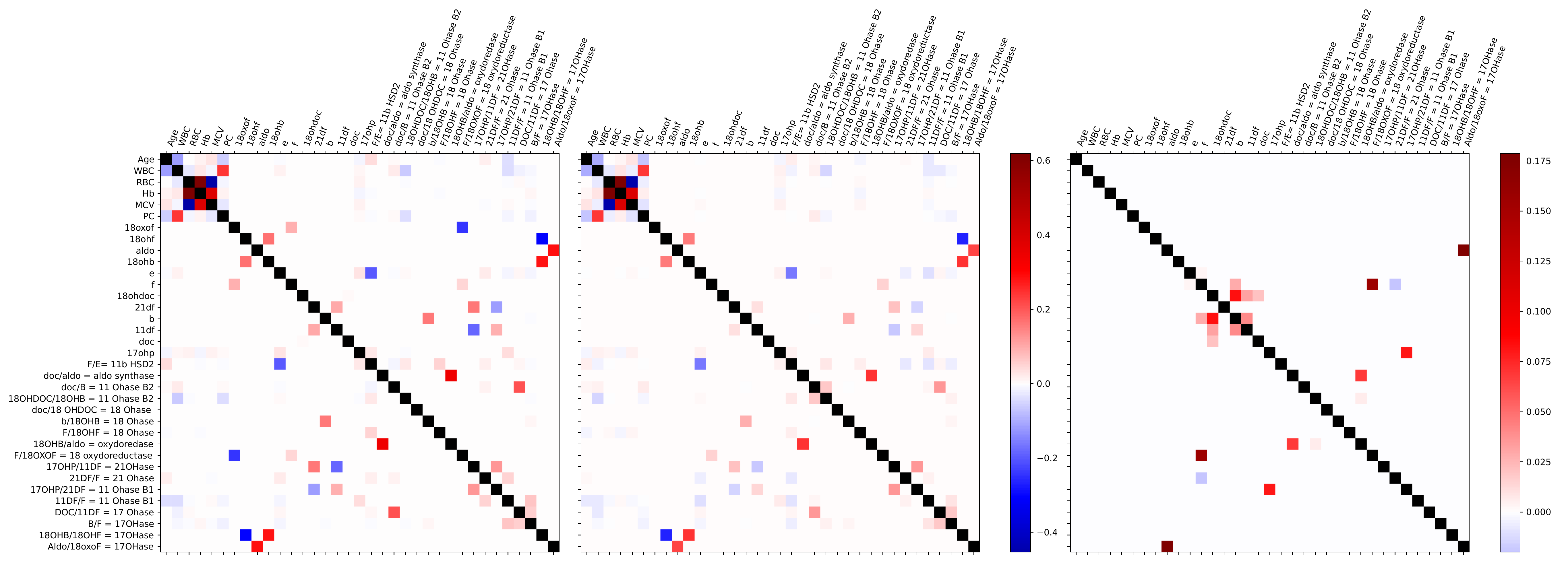}
    \caption{Conditional covariance matrix between the 35 variables measured on the cohort. The positive correlations are in red and the negative in blue. The diagonal coefficients are ignored in this study. GGMselect (left) and Composite (middle) share the same colour scale. The rightmost figure corresponds to one of the sparsest GLASSO solution.}
    \label{fig:cardiovascular_covariance}
\end{figure*}

In this Section, we compare qualitatively the methods in an environment with $p<n$. Although the Composite procedure was developed specifically for the case $n<p$, we demonstrate here that it still holds up to the state of the art outside of its intended application framework. We use a dataset of variables relevant to the adrenal steroidogenesis on a cohort of healthy test subjects.\\
Adrenal steroid synthesis in childhood is a complex process involving an enzymatic cascade that transforms cholesterol into mineralocorticoids, glucocorticoids or androgens, depending on the enzymatic equipment of each zona of the adrenal gland. Even though most important ways of adrenal steroidogenesis are known, we now assess new related metabolite that may ask new questions regarding adrenal steroidogenesis. 
Thus, we analysed a pediatric cohort of $n = 172$ healthy volunteers aged from 3 months to 16 years old with blood count and LC-MS/MS adrenal steroid profile analysis ($p = 35$). \\
\\
\figurename~\ref{fig:cardiovascular_covariance} represents the matrices of pairwise conditional correlations corresponding to the GGMselect solution (left), the Composite solution (middle) and a sparse GLASSO solution (right). The rest of the path of GLASSO solution can be found in the supplementary materials. The other solutions contain many more edges than any of the three matrices here.\\
The models proposed by the three matrices have been compared to literature data for hematological parameters and steroidogenesis analysis. Regarding hematological analysis, both the Composite and GGMselect models confirm well known relations such as strong direct positive links between hemoglobin concentration (Hb) and red cells count (RBC); between hemoglobin concentration and mean corpuscular volume (WCV); between white cells (WBC) and platelet counts (PC); and a strong negative link between red cells count and mean corpuscular volume; between white cells count and age. The GLASSO solution did not show any of them.\\
Regarding steroid metabolism, 11-$\beta$1 hydroxylase (11 Ohase B1) and 21 hydroxylase (21 Ohase) activities, the Composite method and GGMselect reach the same conclusion: there is a strong positive direct link between enzymatic activities and the concentration of their corresponding alternate product. This is in accordance with common description of adrenal steroidogenesis process: decreased activity leads to an accumulation product of the alternative pathway. The GLASSO solution failed to show these relations. In the same way, GGMselect and the Composite method exhibit a negative link between the lack of 11-$\beta$ HSD type 2 (11b HSD2) activity (that catabolizes cortisol into cortisone) and the concentration of its product, cortisone (e). The sparse GLASSO fails to underline this link. All these data tend to show a better interpretation of steroids profile with the GGMselect and Composite solutions. Interestingly, these models also underline a new link: a strong positive link between 18-hydroxycorticosterone (18ohb) and 18-hydroxycortisol (18ohf) concentrations, two steroids that are supposed to be independently produced in two different zonas of the adrenal gland. This result could imply an alternative pathway in adrenal steroidogenesis that needs to be explored.\\
The GGMselect and Composite graphs are mostly identical, although some of the conditional correlations are weaker in the Composite matrix. Among the subtle differences, two edges that are coherent with the state of the art, and are present in the GGMselect graph, were alleviated in the Composite matrix (resulting in invisible connections in \figurename~\ref{fig:cardiovascular_covariance}): the link between the 18-oxocortisol (18oxof) and cortisol (f) concentrations, and the very strong negative link between the ratio cortisol/18-oxocortisol (F/18oxof) and 18-oxocortisol. The other very few additions and removals in the Composite model are hard to validate or disprove with the current state of the art.\\  
From a medical analysis point of view, all these results are preliminary and will have to be confirmed by more in depth studies. From a purely machine learning point of view, this example illustrates that the Composite method behaves appropriately when $p<n$. In this example, the GGMselect solution seems already acceptable, and the Composite procedure does not deviate too much from it. \\
\\
To summarise these experimental studies, Section \ref{section:Alzheimer} showed the quantitative and qualitative improvements made by the Composite method on real data, in the High Dimension Low Sample size setting ($n < p$) the method was designed for. In this Section, with enough data available ($p < n$), hence outside the intended area of application, the qualitative analysis suggests that, running the Composite procedure does not provide additional benefits, but does not cause any loss either.

\section{Conclusion}
When it came to inferring conditional covariance graphs from a small number of observations, we were dissatisfied with the state of the art GGM methods. In this paper, we quantified the shortcomings in terms of goodness of fit, distribution reconstruction and interpretability of the local approach of \cite{meinshausen2006high} and the global optimisation problem of \cite{yuan2007model, banerjee2008model}. \textit{We proposed a method composed of a structure learning algorithm coupled with model selection criterion}. In the latter, the structure learning steps are a variation of the parallel nodewise linear regressions of \cite{meinshausen2006high} and the model selection steps guided by out of sample versions of the likelihood optimised in \cite{yuan2007model} and \cite{banerjee2008model}. The validity of our method was demonstrated on synthetic and real data when $n < p$. Quantitatively, it consistently reached consequently lower KL divergences and better sparsistency than the aforementioned state of the art paradigms. A qualitative analysis on a neurological data set of real data, revealed that it better recovered the known dynamics of the field. An additional real data experiment, with $p < n$, suggested that the method did not cause any loss when used outside the intended scope of application. In the future, optimising the numerical scheme will allow us to make further quantitative improvements. Such as lower execution times and better performances with less reliance on the initialisation.

\section{Proofs of the main results} \label{appendix:proofs}

\subsection{Basic Cross Entropy calculus for Gaussian vectors} \label{appendix:CE}
In this Section, we offer details and commentary on the Cross Entropy manipulation with normal distributions and prove \eqref{eq_gaussian_ce} and \eqref{eq_gaussian_likelihood}.\\
The formula of the Cross Entropy $\crossent{p}{q}$ is given by:
\begin{equation*} 
        H(p,q) := - \mathbb{E}_p\brack{\text{log}\, q(X)} = \int_x -p(x) ln(q(x)) \mu(\mathrm{d}x) \, .
\end{equation*}
The likelihood $p_{\theta}$ of a parametric distribution $f_{\theta}$ with iid observations $(X^{(1)}, ..., X^{(n)})$ is given by:
\begin{equation*}
    p_{\theta}(X^{(1))}, ..., X^{(n)}) = \prod_{i=1}^n f_{\theta}(X^{(i))})\, .
\end{equation*}
Let  $\hat{f}_n = \frac{1}{n} \sum_{i=1}^n \mathds{1}_{x=X^{(i)}}$ be the empirical distribution of the sample $(X^{(1)}, ..., X^{(n)})$, we see the connection between CE and likelihood:
\begin{equation*}
    \crossent{\hat{f}_n}{f_{\theta}} = - \frac{1}{n}\sum_{i=1}^n \text{log} (f_{\theta}(X_i)) = - \frac{1}{n} \text{log}\, p_{\theta}(x_1, ..., x_n)\, .
\end{equation*}
\begin{IEEEproof}[Proof of \eqref{eq_gaussian_ce} and \eqref{eq_gaussian_likelihood} ]
    In the case of Centered Multivariate Gaussians, let $\crossent{\Sigma_1}{\Sigma_2} := \crossent{f_{\Sigma_1}}{f_{\Sigma_2}}$ and let us omit the constant $\frac{p}{2} ln(2 \pi)$ from the calculations:
    \begin{equation*} 
        \begin{split}
               \crossent{\Sigma_1}{\Sigma_2} &\equiv \int_X f_{\Sigma_1}(x) \parent{- \frac{1}{2} ln(\det{K_2}) + \frac{1}{2} X^T K_2 X} \mathrm{d}X \\
               &= - \frac{1}{2} ln(\det{K_2}) + \frac{1}{2}\int_X f_{\Sigma_1}(x)   \dotprod{XX^T, K_2 } \mathrm{d}X \\
               &= - \frac{1}{2} ln(\det{K_2}) + \frac{1}{2}  \dotprod{\int_X f_{\Sigma_1}(x)  XX^T \mathrm{d}X, K_2}   \\
               &= - \frac{1}{2} ln(\det{K_2}) + \frac{1}{2} \dotprod{\Sigma_1, K_2}\, .
        \end{split}
    \end{equation*}
    In the end, we get \eqref{eq_gaussian_ce}:
    \begin{equation*}
    \boxed{
        \crossent{\Sigma_1}{\Sigma_2} \equiv \frac{1}{2} \parent{\dotprod{\Sigma_1, K_2} - ln(\det{K_2})}\, .
        }
    \end{equation*}
    With the observed data  $\underline{X} := \parent{X_1, ..., X_n}^T \in \mathbb{R}^{n \times p}$, let $S := \frac{1}{n} \underline{X}\, \underline{X}^T \in S_p^+$, the empirical covariance matrix. The log likelihood of any centred Gaussian distribution $f_{\Sigma_2}$ is given by:
    \begin{equation*} 
        \begin{split}
               \crossent{\hat{f}_n}{f_{\Sigma_2}} &\equiv \frac{1}{2n} \sum_{i=1}^n \parent{-  ln(\det{K_2}) +  X_i^T K_2 X_i}\\
               &= - \frac{1}{2} ln(\det{K_2}) + \dotprod{\sum_{i=1}^n \frac{ X_i X_i^T}{2n}, K_2}\\
               &= - \frac{1}{2} ln(\det{K_2}) + \frac{1}{2} \dotprod{S, K_2}\, ,
        \end{split}
    \end{equation*}
    where, as in \eqref{eq_gaussian_ce}, we omit the constant term $\frac{p}{2} ln(2 \pi)$ from the calculations.
    In the end, we get \eqref{eq_gaussian_likelihood}:
        \begin{equation*}
    \boxed{
        \crossent{\hat{f}_n}{f_{\Sigma_2}} \equiv \frac{1}{2} \parent{\dotprod{S, K_2} - ln(\det{K_2})}\, .
        }
    \end{equation*}
\end{IEEEproof}
The likelihood $\crossent{\hat{f}_n}{f_{\Sigma_2}}$ follows a similar formula as the Cross Entropy between two normal distributions (\ref{eq_gaussian_ce}). When $S$ defines a non degenerate normal distribution, what we actually have is $\crossent{\hat{f}_n}{f_{\Sigma_2}} = \crossent{f_{S} }{f_{\Sigma_2}}$. However, when $n<p$, $S$ is singular and the density $f_{S} $ is not defined. The formula (\ref{eq_gaussian_likelihood}) still holds though, and we write $\crossent{S}{\Sigma_2} := \crossent{\hat{f}_{n}}{f_{\Sigma_2}}$ since the formula is the same as (\ref{eq_gaussian_ce}) for $\crossent{\Sigma_1}{\Sigma_2}$.
\begin{remark}
    When the density $f_{S} $ does exists, we have equality in the CE $\crossent{\hat{f}_n}{f_{\Sigma_2}} = \crossent{f_{S} }{f_{\Sigma_2}}$, but not in the Entropies $\crossent{\hat{f}_n}{\hat{f}_n} \neq \crossent{f_{S} }{f_{S} }$, as a consequence the KL divergences are different as well: $KL\parent{\hat{f}_n, f_{\Sigma_2}} \neq KL\parent{f_{S} , f_{\Sigma_2}}$. In practice $KL\parent{f_{S} , f_{\Sigma_2}} << KL\parent{\hat{f}_n, f_{\Sigma_2}}$ and $KL\parent{\hat{f}_n, f_{\Sigma_2}}$ will never reach 0, since a normal distribution will tend to be closer to another normal distribution than to an empirical one, this is particularly true with $n$ small and $\Sigma_2$ close to $S$. 
    As a result, $KL\parent{\hat{f}_n, f_{\Sigma_2}}$ offers a poor sense of scale, since the value 0 cannot be used as a reference. For this reason, when we represent $\crossent{f_{S_{test}}}{f_{\Sigma_2}}$ as we do in \figurename~\ref{fig:real_KL_sparsity_glasso}, we do not use it under the form of a KL with 0 as its minimum for scale reference - as we do on synthetic data in \figurename~\ref{fig:grid_KL} - since the only KL we can compute is the mostly irrelevant $KL\parent{\hat{f}_n, f_{\Sigma_2}}$.
\end{remark}

\subsection{Preliminary results for the model selection guarantees}
To prove the controls we stated in Sections \ref{section:th_basic}, \ref{section:th_results_expectation} and \ref{section:th_results_probability}, we need the two following lemmas.
\begin{lemma}
    \textit{Let $S^{(\lambda)} := S + \lambda I_p$. With $\widehat{K}_{\G} := \widehat{\Sigma}_{\G}^{-1}$, where $\widehat{\Sigma}_{\G}$ is defined as in (\ref{eq_def_max_vrais_lambda}), we have:}
    \begin{equation} \label{eq_dotprod_mle}
        \forall \G \in \M, \quad \dotprod{S^{(\lambda)}, \widehat{K}_{\G}} = p\, .
    \end{equation}
\end{lemma}
\begin{IEEEproof}
    Let $\Pi_{\G}$ be the orthogonal projection on the edge set $E_{\G} \cup \brace{(i,i)}_{i=1}^p$. That is to say, for any matrix $M\in \R^{p \times p}, \quad \Pi_{\G}(M)_{i,j} = M_{i,j}  \mathds{1}_{ (i, j) \in E_{\G} \cup \brace{(i,i)}_{i=1}^p}$. A property of the MLE is that $\Pi_{\G}(\widehat{\Sigma}_{\G}) = \Pi_{\G}(S^{(\lambda)})$, i.e. the matrices have the same values on the diagonal and the edge set, see \cite{dempster1972covariance}. Additionally, note that, because of the sparsity of $\widehat{K}_{\G}$, for any matrix $M$, we have $\dotprod{M, \widehat{K}_{\G}} = \dotprod{\Pi_{\G}(M), \widehat{K}_{\G}}$. Then:
    \begin{equation*}
    \begin{split}
        \dotprod{S^{(\lambda)}, \widehat{K}_{\G}} &= \dotprod{\Pi_{\G}(S^{(\lambda)}), \widehat{K}_{\G}} \\
        \dotprod{S^{(\lambda)}, \widehat{K}_{\G}} &= \dotprod{\Pi_{\G}(\widehat{\Sigma}_{\G}), \widehat{K}_{\G}}\\
        \dotprod{S^{(\lambda)}, \widehat{K}_{\G}} &= \dotprod{\widehat{\Sigma}_{\G}, \widehat{K}_{\G}}\\
        \dotprod{S^{(\lambda)}, \widehat{K}_{\G}} &= p\,.
    \end{split}
    \end{equation*}
\end{IEEEproof}

\begin{lemma}
    \textit{With $\widehat{K}_{\G}:=\widehat{\Sigma}_{\G}^{-1}$, where $\widehat{\Sigma}_{\G}$ is defined as (\ref{eq_def_max_vrais_lambda}), we have:}
    \begin{equation*}
        \norm{\widehat{K}_{\G}}_* \leq \frac{p}{\lambda}\,.
    \end{equation*}
\end{lemma}
\begin{IEEEproof}
We have:
    \begin{equation*}
    \begin{split}
        \dotprod{S + \lambda I_p, \widehat{K}_{\G}} &= p\\
        \dotprod{S, \widehat{K}_{\G}} + \lambda tr(\widehat{K}_{\G}) &= p\\
        tr\parent{\widehat{K}_{\G}^{\frac{1}{2}}  S \widehat{K}_{\G}^{\frac{1}{2}}}+ \lambda tr(\widehat{K}_{\G}) &= p\,.
    \end{split}
    \end{equation*}
    Since $\widehat{K}_{\G}^{\frac{1}{2}}  S \widehat{K}_{\G}^{\frac{1}{2}} \in S_p^+$, we have $tr\parent{\widehat{K}_{\G}^{\frac{1}{2}}  S \widehat{K}_{\G}^{\frac{1}{2}}} \geq 0 $ and $\lambda tr(\widehat{K}_{\G}) \leq p$, i.e. 
    \begin{equation*}
        \norm{\widehat{K}_{\G}}_* \leq \frac{p}{\lambda}\,.
    \end{equation*}
\end{IEEEproof}

\subsection{Bounds in expectation for the CVCE solutions
}
We prove the results of Sections \ref{section:th_basic} and \ref{section:th_results_expectation}.\\
\begin{IEEEproof}[Proof of \eqref{eq_expectation_cv_guarantee}, \eqref{eq_cv_control_order}, \eqref{eq_cv_control_order2}, and \eqref{eq_cv_control_lounici} 
] 
We want to control the expected regret $e := \E{\crossent{\Sigma}{\widehat{\Sigma}_{\widehat{\G}_{CV}}} - \crossent{\Sigma}{\widehat{\Sigma}_{\widehat{\G}^*}}}$. First, note that by definition of $\widehat{\G}^*$, we have
\begin{equation*}
    0 \leq \crossent{\Sigma}{\widehat{\Sigma}_{\widehat{\G}_{CV}}} - \crossent{\Sigma}{\widehat{\Sigma}_{\widehat{\G}^*}}\,.
\end{equation*}
So the lower bound:
\begin{equation*}
    0 \leq e \, ,
\end{equation*}
is guaranteed.\\
From the definition of $\widehat{\G}_{CV}$ (\ref{eq_model_cv}), we get:
\begin{equation*}
    \crossent{S_{val}}{\widehat{\Sigma}_{\widehat{\G}_{CV}}} \leq \crossent{S_{val}}{\widehat{\Sigma}_{\widehat{\G}^*}}\,.
\end{equation*}
We have for any $\widetilde{\Sigma} \in S_p^{++}$, with $\widetilde{K} := \widetilde{\Sigma}^{-1}$:
\begin{equation*}
    \crossent{S_{val}}{\widetilde{\Sigma}} =  \crossent{\Sigma}{\widetilde{\Sigma}} + \frac{1}{2} \dotprod{S_{val} - \Sigma, \widetilde{K}}\,.
\end{equation*}
Hence:
\begin{equation} \label{eq_cv_inequality}
\begin{split}
        \crossent{\Sigma}{\widehat{\Sigma}_{\widehat{\G}_{CV}}} \leq \crossent{\Sigma}{\widehat{\Sigma}_{\widehat{\G}^*}} &+ \frac{1}{2} \dotprod{S_{val} - \Sigma, \widehat{K}_{\widehat{\G}^*}} \\
        &- \frac{1}{2} \dotprod{S_{val} - \Sigma, \widehat{K}_{\widehat{\G}_{CV}}}\,.
\end{split}
\end{equation}
Since $K_{\widehat{\G}^*}$ is defined from $S_{expl}$ uniquely, and independently of $S_{val}$, we get
\begin{equation} \label{eq_exp_null}
\begin{split}
        \E{\dotprod{S_{val} - \Sigma, \widehat{K}_{\widehat{\G}^*}} \Big| S_{expl}} &= \dotprod{\E{S_{val} - \Sigma | S_{expl}}, \widehat{K}_{\widehat{\G}^*}}\\
        &= 0\,.
\end{split}
\end{equation}
From (\ref{eq_cv_inequality}) and (\ref{eq_exp_null}) we get:
\begin{equation*} 
\begin{split}
    \E{\crossent{\Sigma}{\widehat{\Sigma}_{\widehat{\G}_{CV}}}} \leq\:&  \E{\crossent{\Sigma}{\widehat{\Sigma}_{\widehat{\G}^*}}} \\
    &+ \frac{1}{2}\E{  \dotprod{\Sigma - S_{val}, \widehat{K}_{\widehat{\G}_{CV}}}}\,.
\end{split}
\end{equation*}
Which is exactly the result of Eq.~\eqref{eq_expectation_cv_guarantee}:
\begin{equation*} 
    \boxed{
e \leq \frac{1}{2} \E{\dotprod{\Sigma-S_{val}, \widehat{K}_{\widehat{\G}_{CV}}}}\, .
}
\end{equation*}
As we discussed in Section \ref{section:th_results_expectation}, to obtain Eq.~\eqref{eq_expectation_cv_guarantee}, we only used the definitions of $\widehat{\G}_{CV}$ for the upper bound and $\widehat{\G}^*$ for the lower bound. Since we assume nothing on the model family $\M$, those bounds are somewhat optimal in terms of the available information. Additionally, (\ref{eq_expectation_cv_guarantee}) is actually independent of how the symmmetric positive matrices $\{\widehat{\Sigma}_{\G}\}_{\G \in \M}$ are defined as long as they are function only of $S_{expl}$. They do not need to be associated with a different graph each, or with any graph for that matter. They do not need to be solutions of the MLE problem (\ref{eq_def_max_vrais_lambda}) and could be for example all the solutions on the path of solution of the $l_1-$penalised likelihood optimisation problem \eqref{eq_likelihood_l1}.\\
To get a more explicit control on the CVCE however, we need the assumption that $\widehat{\Sigma}_{\G}$ is the constrained MLE defined in (\ref{eq_def_max_vrais_lambda}).\\
Let $\Sigma_{\infty} := \max{i,j}\det{\Sigma_{ij}}$. We call $E_{\text{max}}$ the union of the maximal edge sets in $\M$, $d_{\text{max}} = \det{E_{max}} \leq \frac{p(p-1)}{2}$ its cardinal and $\Pi_{max}$ the orthogonal projection on $E_{\text{max}}\cup \brace{(i,i)}_{i=1}^p$. We have:

\begin{equation*}
\begin{split}
e &\leq \frac{1}{2} \E{\dotprod{\Sigma-S_{val}, \widehat{K}_{\widehat{\G}_{CV}}}} \\
&= \frac{1}{2} \E{\dotprod{\Pi_{\widehat{\G}_{CV}}\parent{\Sigma-S_{val}}, \widehat{K}_{\widehat{\G}_{CV}}}} \\
&\leq \frac{1}{2}\E{\norm{\Pi_{\widehat{\G}_{CV}}\parent{\Sigma-S_{val}}}_F^2}^{\frac{1}{2}} \E{\norm{\widehat{K}_{\widehat{\G}_{CV}}}_F^2}^{\frac{1}{2}} \\
&\leq \frac{1}{2} \E{\norm{\Pi_{max}\parent{\Sigma-S_{val}}}_F^2}^{\frac{1}{2}} \E{\norm{\widehat{K}_{\widehat{\G}_{CV}}}_*^2}^{\frac{1}{2}}\\
&\leq \frac{1}{2} \Bigg(\sum_{i = 1}^p \E{\parent{\Sigma^{ii}-S^{ii}_{val}}^2} \\
& \hspace{1cm}+ \sum_{(i, j) \in E_{max}}\E{\parent{\Sigma^{ij}-S^{ij}_{val}}^2}\Bigg)^{\frac{1}{2}} \frac{p}{\lambda}\\
&\leq \frac{1}{2}  \parent{ \frac{2 \Sigma_{\infty}^2}{n_{val}} \parent{p + 2 d_{max}}}^{\frac{1}{2}} \frac{p}{\lambda}\,.
\end{split}
\end{equation*}
From which we finally get the result of \eqref{eq_cv_control_order}:

\begin{equation*} 
    \boxed{
e \leq \frac{\Sigma_{\infty}}{\lambda \sqrt{2}} \frac{\parent{p + 2 d_{max}}^{\frac{1}{2}} p}{\sqrt{n_{val}}} \, .
     }
\end{equation*}
If $E_{\text{max}}$ is dependent on the \textit{exploration} data - because the graph family $\M$ was built from $S_{expl}$ for instance - we have:

\begin{alignat*}{3}
    \E{\norm{\Pi_{max}\parent{\Sigma-S_{val}}}_F^2}^{\frac{1}{2}} & \quad \: &&\\
    &\hspace{-3cm} = &&\hspace{-3cm} \Bigg(\sum_{i = 1}^p \E{\parent{\Sigma^{ii}-S^{ii}_{val}}^2}\\
    &\hspace{-3cm}  &&\hspace{-3cm} + \E{\sum_{i, j \in E_{max}}\E{\parent{\Sigma^{ij}-S^{ij}_{val}}^2 \Big| S_{expl}}}\Bigg)^{\frac{1}{2}} \\
    &\hspace{-3cm} \leq &&\hspace{-3cm} \parent{ \frac{2 \Sigma_{\infty}^2}{n_{val}} \parent{p + 2 \E{d_{max}}}}^{\frac{1}{2}}\, .
\end{alignat*}
We get the control \eqref{eq_cv_control_order2}, the same as \eqref{eq_cv_control_order} but with an additional expectation term:
\begin{equation*} 
    \boxed{
    \begin{aligned}
    e \leq &\frac{\Sigma_{\infty}}{\lambda \sqrt{2}} \frac{\parent{p + 2 \E{d_{max}}}^{\frac{1}{2}} p}{\sqrt{n_{val}}} \, .
    \end{aligned}
     }
\end{equation*}
In order to prove \eqref{eq_cv_control_lounici}, 
we start by showing how the regret is bounded by operator norm $\norm{\Sigma-S_{val}}_2$. By tracial matrix Holder inequality:
\begin{equation*} 
\begin{split}
    \dotprod{S_{val} - \Sigma, \widehat{K}_{\widehat{\G}_{CV}}} &\leq \norm{\Sigma-S_{val}}_2 \norm{\widehat{K}_{\widehat{\G}_{CV}}}_* \\
    &= \norm{\Sigma-S_{val}}_2 tr\parent{\widehat{K}_{\widehat{\G}_{CV}}}\\
    &\leq \frac{\norm{\Sigma-S_{val}}_2}{\lambda} p\, .
\end{split}
\end{equation*}
Then, using \eqref{eq_expectation_cv_guarantee}, we get:
\begin{equation} \label{eq:control_operator_norm}
    e \leq  \E{\norm{\Sigma-S_{val}}_2} \frac{p}{2\lambda} \, .
\end{equation}
To prove \eqref{eq_cv_control_lounici}, we first recall Theorem 4 of \cite{koltchinskii2014concentration}:\\
\\
\textbf{Theorem 4 of \cite{koltchinskii2014concentration}.}
    \textit{Let $X_1, X_2, ..., X_n$ be i.i.d. weakly square integrable centered random vectors in a separable Banach space with norm $\norm{.}$ and $\Sigma$ be their covariance operator. If X is Gaussian, then there exist an absolute constant $c$, independent of the problem, such that:}
    \begin{equation} \label{eq:thm_koltchinskii_lounici}
        \E{\norm{\widehat{\Sigma} - \Sigma}} \leq c \norm{\Sigma} max\parent{\sqrt{\frac{\E{\norm{X}}^2}{n \norm{\Sigma}}}, \frac{\E{\norm{X}}^2}{n \norm{\Sigma}}},
    \end{equation}
     \textit{where $\norm{.}$ for operators denotes the operator norm associated with the vector norm $\norm{.}$, that is to say:}
     \begin{equation*}
         \norm{\Sigma} = \underset{\norm{u} = 1}{sup} \, \norm{\Sigma u} \,.
     \end{equation*}
In our case, $X \sim \N{0_p, \Sigma}$ is a Gaussian vector in the Banach space $\R^p$, with the euclidean norm $\norm{X}_2$, that verifies the integrability properties of the Theorem and whose covariance operator is the covariance matrix $\Sigma$. Hence the theorem can be applied. The operator norm for a symmetric positive matrix $\Sigma$ associated with the euclidean norm is also called the spectral norm, since it corresponds to the highest eigenvalue: $\norm{\Sigma}_2 = \lambda_{max}(\Sigma)$.\\
For a Gaussian vector: $Z \sim \N{0_p, I_p}$, we have:
\begin{equation*}
    \E{\norm{Z}_2} \leq \sqrt{p}.
\end{equation*}
Since $K^{\frac{1}{2}} X \sim \N{0_p, I_p}$, and 
\begin{equation*}
    \begin{split}
        \norm{X}_2 &= \norm{\Sigma^{\frac{1}{2}} K^{\frac{1}{2}} X}_2 \\
        &\leq  \norm{\Sigma^{\frac{1}{2}}}_2 \norm{K^{\frac{1}{2}} X}_2 \, ,
    \end{split}
\end{equation*}
we have:
\begin{equation*}
    \E{\norm{X}_2} \leq \norm{\Sigma^{\frac{1}{2}}}_2 \sqrt{p} \, .
\end{equation*}
Since $\norm{\Sigma}_2 = \lambda_{max}(\Sigma)$, we have by definition, $\norm{\Sigma^{\frac{1}{2}}}_2 =\norm{\Sigma}_2^{\frac{1}{2}}$. In the end, when we apply \eqref{eq:thm_koltchinskii_lounici} to our case, we get:
\begin{equation}
    \E{\norm{S_{val} - \Sigma}_2} \leq c \lambda_{max}(\Sigma) max\parent{ \sqrt{\frac{p}{n_{val}}}, \frac{p}{n_{val}} } \, .
\end{equation}
We apply this concentration result on \eqref{eq:control_operator_norm} to obtain \eqref{eq_cv_control_lounici}:
\begin{equation*}
\boxed{
e \leq  c \frac{\lambda_{max}(\Sigma)}{\lambda} p \parent{\sqrt{\frac{p}{n_{val}}} \vee \frac{p}{n_{val}}}\,.
}
\end{equation*}


\end{IEEEproof}

\subsection{Bounds in probability for the CVCE solutions
}
We prove the results of Section \ref{section:th_results_probability}.\\
\begin{IEEEproof} [Proof of \eqref{eq_concentration_1} and \eqref{eq_concentration_2}]
We want to lower bound the probability that the regret is small: $P := \P{\det{\crossent{\Sigma}{\widehat{\Sigma}_{\widehat{\G}_{CV}}} - \crossent{\Sigma}{\widehat{\Sigma}_{\widehat{\G}^*}}}\leq \delta}$. The concentration dynamic driving the results comes from the convergence of random Wishart matrix $S_{val}$ towrds its average $\Sigma$, which is made stronger by the number of observations $n_{val}$ in the \textit{validation} set. Since: 
\begin{equation*}
\begin{split}
        \det{\crossent{\Sigma}{\widehat{\Sigma}_{\widehat{\G}_{CV}}} - \crossent{\Sigma}{\widehat{\Sigma}_{\widehat{\G}^*}}} &\leq \\
        &\hspace{-0.7cm} \det{\crossent{\Sigma}{\widehat{\Sigma}_{\widehat{\G}_{CV}}} - \crossent{S_{val}}{\widehat{\Sigma}_{\widehat{\G}_{CV}}}} \\
        &\hspace{-0.7cm} +  \det{\crossent{S_{val}}{\widehat{\Sigma}_{\widehat{\G}^*}} - \crossent{\Sigma}{\widehat{\Sigma}_{\widehat{\G}^*}}},
\end{split}
\end{equation*}
then
\begin{equation*}
\begin{split}
    &\forall \G \in \M, \; \det{\crossent{S_{val}}{\widehat{\Sigma}_{\G}} - \crossent{\Sigma}{\widehat{\Sigma}_{\G}}} \leq \frac{\delta}{2}\\ 
    \implies  &\det{\crossent{\Sigma}{\widehat{\Sigma}_{\widehat{\G}_{CV}}} - \crossent{\Sigma}{\widehat{\Sigma}_{\widehat{\G}^*}}}\leq \delta\,.
\end{split}
\end{equation*}
Since:
\begin{equation*}
    \crossent{S_{val}}{\widehat{\Sigma}_{\G}} - \crossent{\Sigma}{\widehat{\Sigma}_{\G}} = \frac{1}{2} \dotprod{S_{val} - \Sigma, \widehat{K}_{\G}} \, ,
\end{equation*}
then
\begin{equation} \label{eq_ineq_inter_concentration_2}
\begin{split}
    &\forall \G \in \M, \; \det{\dotprod{S_{val} - \Sigma, \widehat{K}_{\G}}} \leq \delta\\ 
    \implies  &\det{\crossent{\Sigma}{\widehat{\Sigma}_{\widehat{\G}_{CV}}} - \crossent{\Sigma}{\widehat{\Sigma}_{\widehat{\G}^*}}}\leq \delta\,.
\end{split}
\end{equation}

From the 
logical implication (\ref{eq_ineq_inter_concentration_2}), we can take two path to derive two different bounds: one with a more general expression, and a more precise one taking into consideration the sparsity of the models. For the first one, note that $S_{val} = \Sigma^{\frac{1}{2}} W \Sigma^{\frac{1}{2}}$ where $n_{val}W \sim \W{I_p, n_{val}}{p}$ is a standard Wishart matrix. Then we have:
\begin{equation*}
\begin{split}
        \forall \G, \; \dotprod{S_{val} - \Sigma, \widehat{K}_{\G}} &= \dotprod{W - I_p, \Sigma^{-\frac{1}{2}}\widehat{K}_{\G}\Sigma^{-\frac{1}{2}}}\\ 
    &\leq \norm{W-I_p}_F \norm{\Sigma^{-\frac{1}{2}}\widehat{K}_{\G}\Sigma^{-\frac{1}{2}}}_F\\
    &\leq \norm{W-I_p}_F \max{\G \in \M}\norm{\Sigma^{-\frac{1}{2}}\widehat{K}_{\G}\Sigma^{-\frac{1}{2}}}_F\,.
\end{split}
\end{equation*}
We plug this result into 
\eqref{eq_ineq_inter_concentration_2} to obtain:
\begin{equation*}
    \begin{split}
        &\norm{W-I_p}_F  \max{\G \in \M}\norm{\Sigma^{-\frac{1}{2}}\widehat{K}_{\G}\Sigma^{-\frac{1}{2}}}_F \leq \delta \\
        \implies &\forall \G \in \M, \; \dotprod{S_{val} - \Sigma, \widehat{K}_{\G}} \leq \delta \\
        \implies &\det{\crossent{\Sigma}{\widehat{\Sigma}_{\widehat{\G}_{CV}}} - \crossent{\Sigma}{\widehat{\Sigma}_{\widehat{\G}^*}}}\leq \delta\, .
    \end{split}
\end{equation*}
We end up with the control \eqref{eq_concentration_1} by taking the probability in the previous expression: 
\begin{equation*} 
\boxed{
P \geq \P{\norm{W-I_p}_F  \leq \frac{\delta}{\max{\G \in \M}\norm{\Sigma^{-\frac{1}{2}}\widehat{K}_{\G}\Sigma^{-\frac{1}{2}}}_F}}\, .
}
\end{equation*}
For the second result, let $\Pi_{\G}$ and $\Pi_{max}$ be the orthogonal projections on the edge sets $E_{\G}\cup \brace{(i,i)}_{i=1}^p$ and $E_{max}\cup \brace{(i,i)}_{i=1}^p$ respectively. We have:
\begin{equation*}
\begin{split}
        \forall \G, \; \dotprod{S_{val} - \Sigma, \widehat{K}_{\G}} &= \dotprod{\Pi_{\G}(S_{val} - \Sigma), \widehat{K}_{\G}}\\ 
    &\leq \norm{\Pi_{\G}(S_{val} - \Sigma)}_F \norm{\widehat{K}_{\G}}_F\\
    &\leq \norm{\Pi_{max}(S_{val} - \Sigma)}_F \max{\G \in \M} \norm{\widehat{K}_{\G}}_F\,.
\end{split}
\end{equation*}
Hence we get, from 
\eqref{eq_ineq_inter_concentration_2}, the logical implication:
\begin{equation*}
    \begin{split}
        &\norm{\Pi_{max}(S_{val} - \Sigma)}_F \max{\G \in \M} \norm{\widehat{K}_{\G}}_F\ \leq \delta \\
        \implies &\forall \G, \; \dotprod{S_{val} - \Sigma, \widehat{K}_{\G}} \leq \delta \\
        \implies &\det{\crossent{\Sigma}{\widehat{\Sigma}_{\widehat{\G}_{CV}}} - \crossent{\Sigma}{\widehat{\Sigma}_{\widehat{\G}^*}}}\leq \delta\, .
    \end{split}
\end{equation*}
From which we get the control \eqref{eq_concentration_2} by taking the probability of the events:
\begin{equation*} 
\boxed{
P \geq \P{\norm{\Pi_{max}(S_{val} - \Sigma)}_F \leq \frac{\delta}{\max{\G \in \M}  \norm{\widehat{K}_{\G}}_F}} \,.
}
\end{equation*}
We underline that we obtain the two controls \eqref{eq_concentration_1} and \eqref{eq_concentration_2} directly from logical implications. Hence, they remain true when every probability is taken conditionally to any random variable, for instance the \textit{exploration} data set, or the sufficient statistic built from it: $S_{expl}$.
\end{IEEEproof}

\begin{remark}
    Since $\forall \G \in \M, \; \norm{\widehat{K}_{\G}}_* \leq \frac{p}{\lambda}$, both $\max{\G \in \M}\norm{\Sigma^{-\frac{1}{2}}\widehat{K}_{\G}\Sigma^{-\frac{1}{2}}}_F$ and $\max{\G \in \M} \norm{\widehat{K}_{\G}}_F$ are bounded random variables. They depend only on the \textit{exploration} empirical covariance $S_{expl}$ and can be seen as constants of the problem if working conditionally to the \textit{exploration} set. Likewise, $\Pi_{max}$ is a deterministic function conditionally to $S_{expl}$.
\end{remark}

\bibliographystyle{IEEEtran}
\bibliography{references}

\section*{Acknowledgments}
The research leading to these results has received funding from the European Research Council (ERC) under grant agreement No 678304, European Union’s Horizon 2020 research and innovation program under grant agreement No 666992 (EuroPOND) and No 826421 (TVB-Cloud), and the French government under management of Agence Nationale de la Recherche as part of the "Investissements d'avenir" program, reference ANR-19-P3IA-0001 (PRAIRIE 3IA Institute) and reference ANR-10-IAIHU-06 (IHU-A-ICM). The authors would like to thank Pascal Houillier for his insightful comments on the nephrological experiments.

%

\begin{IEEEbiography}[{\includegraphics[width=1in,height=1.25in,clip,keepaspectratio]{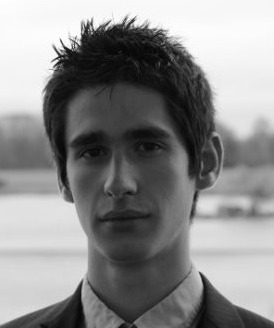}}]{Thomas Lartigue} received graduate degrees from the \'Ecole polytechnique and the \'Ecole Normale Sup\'erieur Paris-Saclay. He is carrying a PhD thesis funded by INRIA at the Centre de Math\'ematiques Appliqu\'ees of Ecole polytechnique. His research interests include Computational Statistics and Machine Learning, ranging from parameter estimation and Bayesian inference to optimisation.  
\end{IEEEbiography}

\begin{IEEEbiography}[{\includegraphics[width=1in,height=1.25in,clip,keepaspectratio]{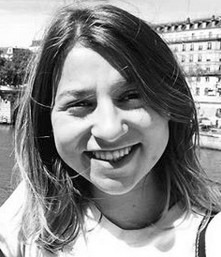}}]{Simona Bottani} is a PhD student funded by INRIA at Aramis Lab team at the Brain and Spine Institute in Paris. Her PhD focuses on machine learning for differential diagnosis of neurodegenerative diseases. Bottani received a master degree on Biomedical engineer at Politecnico di Torino in December 2016.
\end{IEEEbiography}

\begin{IEEEbiography}[{\includegraphics[width=1in,height=1.25in,clip,keepaspectratio]{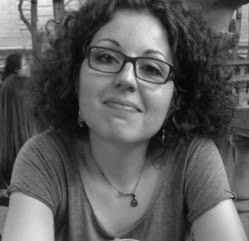}}]{St\'ephanie Baron} is a biochemist at Georges Pompidou European Hospital (Assistance Publique-Hopitaux de Paris), in Physiology Department. Her research interests include endocrinology, adrenal gland and hypertension. She received a PhD degree in Physiology from Paris Descartes University.
\end{IEEEbiography}


\newpage

\begin{IEEEbiography}[{\includegraphics[width=1in,height=1.25in,clip,keepaspectratio]{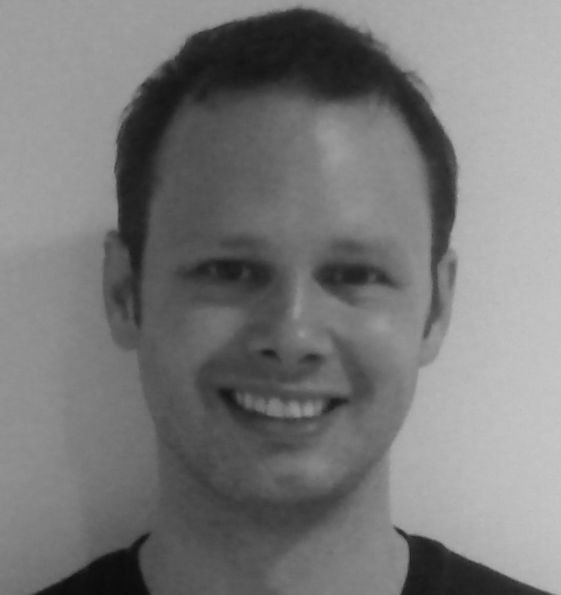}}]{Olivier Colliot}, PhD, is a Research Director at CNRS. He is the co-head of the ARAMIS Lab (Paris, France), a multidisciplinary laboratory dedicated to data science and machine learning applied to neurological diseases. His research interests include medical image computing, machine learning, image analysis and decision support systems. He received the PhD from Telecom ParisTech in 2003 and the Habilitation degree from University Paris-Sud in 2011. He is a member of the Editorial Board of Medical Image Analysis (Elsevier). 
\end{IEEEbiography}

\begin{IEEEbiography}[{\includegraphics[width=1in,height=1.25in,clip,keepaspectratio]{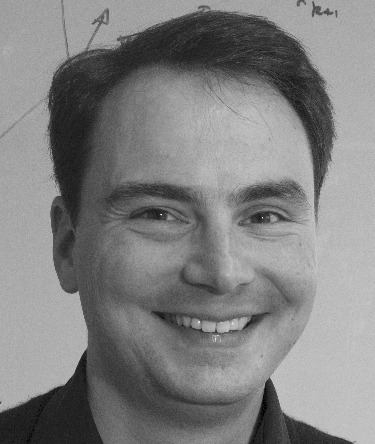}}]{Stanley Durrleman} is INRIA researcher, co-head of the ARAMIS Lab at the Brain Institute in Paris and founding director of the ICM Center for Neuroinformatics. He has developed statistical and computational approaches to create personalised digital brain models from multimodal patients data including image and clinical data. These models reproduce and predict the effect of a disease on brain anatomy and function in any patient. He received several awards including the MICCAI young investigator award and ERC starting grant from the European research council.\end{IEEEbiography}

\begin{IEEEbiography}[{\includegraphics[width=1in,height=1.25in,clip,keepaspectratio]{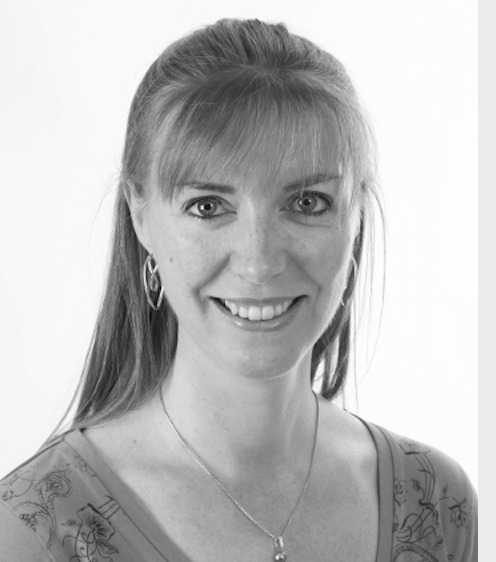}}]{St\'ephanie Allassonni\`ere}, Professor of Applied Mathematics in Paris Descartes School of medicine.
She received her PhD degree in Applied Mathematics (2007), studies one year as postdoctoral fellow in the Center for Imaging Science, JHU, Baltimore. She joined the Applied Mathematics department of Ecole Polytechnique in 2008 as assistant professor and moved to Paris Descartes school of medicine in 2016 as Professor. Her researches focus on statistical analysis of medical databases in order to: understanding the common features of populations, designing classification, early prediction and decision support systems.\end{IEEEbiography}




\end{document}